\documentclass[letterpaper,journal]{IEEEtran}
\usepackage{amsmath,amsfonts}
\usepackage{algorithmic}
\usepackage{algorithm}
\usepackage{array}
\usepackage[caption=false,font=normalsize,labelfont=sf,textfont=sf]{subfig}
\usepackage{textcomp}
\usepackage{pifont}
\usepackage{stfloats}
\usepackage{url}
\usepackage{verbatim}
\usepackage{graphicx}
\usepackage{cite}

\usepackage{booktabs}
\usepackage{multirow}
\usepackage[table]{xcolor}
\usepackage{capt-of}
\usepackage{graphicx}
\usepackage{adjustbox}

\usepackage{graphicx,pgfplots}
\pgfplotsset{compat=1.18}

\usepackage[table]{xcolor}
\usepackage{booktabs, colortbl, array}

\definecolor{gainc}{HTML}{047857}   %
\definecolor{baselinec}{HTML}{40AF8A} %
\definecolor{lossc}{HTML}{B91C1C}   %
\definecolor{rulec}{HTML}{9AA3B2}
\newcommand{\metric}[1]{\textcolor{black!85}{#1}}
\newcommand{\frozenhead}{\textcolor{black}{\ding{100}}}

\definecolor{rank1}{HTML}{BCD3F6}
\definecolor{rank2}{HTML}{CDDFF8}
\definecolor{rank3}{HTML}{DCE9FA}
\definecolor{rank4}{HTML}{EAF2FC}
\definecolor{rank5}{HTML}{F5F9FE}

\newlength{\cw}
\newcolumntype{C}{>{\centering\arraybackslash}p{\cw}}

\newsavebox{\numbox}

\newcommand{\rkx}[4]{\ifnum#1>0 \cellcolor{rank#1}\fi%
  \sbox{\numbox}{\ifnum#1=1 \textbf{#2}\else #2\fi}%
  \leavevmode\vtop{%
    \hbox to \wd\numbox{\usebox{\numbox}}%
    \nointerlineskip\vskip 2.5pt
    \hbox to \wd\numbox{\hfill\scriptsize\ifnum#1=1 \bfseries\fi\textcolor{#3}{#4}%
      \rule[-4.5pt]{0pt}{4.5pt}}}}

\newcommand{\rk}[3]{\rkx{#1}{#2}{gainc}{+#3}}        %
\newcommand{\rl}[3]{\rkx{#1}{#2}{lossc}{$-$#3}}      %

\newcommand{\bs}[2]{\cellcolor{rank#1}#2}
\makeatletter
\newcommand{\bb}[2]{\cellcolor{rank#1}%
  \leavevmode\rlap{\hspace*{-\tabcolsep}%
    \raisebox{-\dp\@arstrutbox}[0pt][0pt]{%
      \textcolor{baselinec}{\rule{\dimexpr\linewidth+2\tabcolsep\relax}{1.2pt}}}}%
  \makebox[\linewidth][c]{#2}}
\makeatother
\newcolumntype{D}{>{\centering\arraybackslash}p{1.65cm}}

\definecolor{secbg}{HTML}{EEF2F7}   %
\definecolor{secfg}{HTML}{334155}   %
\newcommand{\sect}[1]{%
  \rowcolor{secbg}\multicolumn{7}{c}{\textcolor{secfg}{\textbf{#1}}}\\}

\begin{document}

\usetikzlibrary{arrows.meta}

\newcommand{\methodname}{ViT-Up}
\providecommand{\best}[1]{\textbf{#1}}
\providecommand{\second}[1]{\underline{#1}}
\providecommand{\gain}[1]{{\scriptsize #1}}

\title{ViT-Up: Faithful Feature Upsampling for~Vision~Transformers}

\author{Krispin~Wandel, Jingchuan~Wang, and Hesheng~Wang%
\thanks{K. Wandel, J. Wang, and H. Wang are with Shanghai Jiao Tong University, Shanghai, China. Corresponding author: Hesheng Wang.}%
}

\markboth{arXiv preprint}
{Wandel and Wang: ViT-Up: Faithful Feature Upsampling for Vision Transformers}

\maketitle

\begin{abstract}

Vision Transformers (ViTs) have become a dominant architecture for visual representation learning, providing exceptionally strong and broadly reusable backbone features. However, ViTs are commonly operated on relatively small patch-token grids due to the quadratic cost of global self-attention, which creates a persistent bottleneck for dense prediction tasks such as semantic segmentation and depth estimation. This has motivated the development of task-agnostic feature upsamplers.
While recent state-of-the-art methods produce visually sharp dense representations, their reliance on shallow image encoders for guided upsampling can introduce feature leakage, fragmentation, and blur.
We introduce ViT-Up, an implicit feature upsampling framework that replaces external image guidance with layer-wise query construction from intermediate ViT hidden states. This enables feature prediction at arbitrary continuous image coordinates while preserving alignment with the backbone feature space. Training combines multi-scale feature supervision in Stage~1 with offset-guided self-distillation in Stage~2 for increased sharpness and detail. Experiments demonstrate that ViT-Up consistently outperforms state-of-the-art image-guided upsamplers in dense linear probing and semantic correspondence. On DINOv3-S+, the Stage~2 model improves over the strongest evaluated baselines by +3.08 mIoU on Cityscapes and +5.12 PCK@0.10 points on SPair-71k. With the larger DINOv3-B backbone, these gains increase to +3.47 mIoU and +8.79 PCK@0.10 points, demonstrating that ViT-Up scales favorably with backbone capacity.

\end{abstract}


\section{Introduction}

\begin{figure}[!t]
    \centering
    \includegraphics[width=\columnwidth]{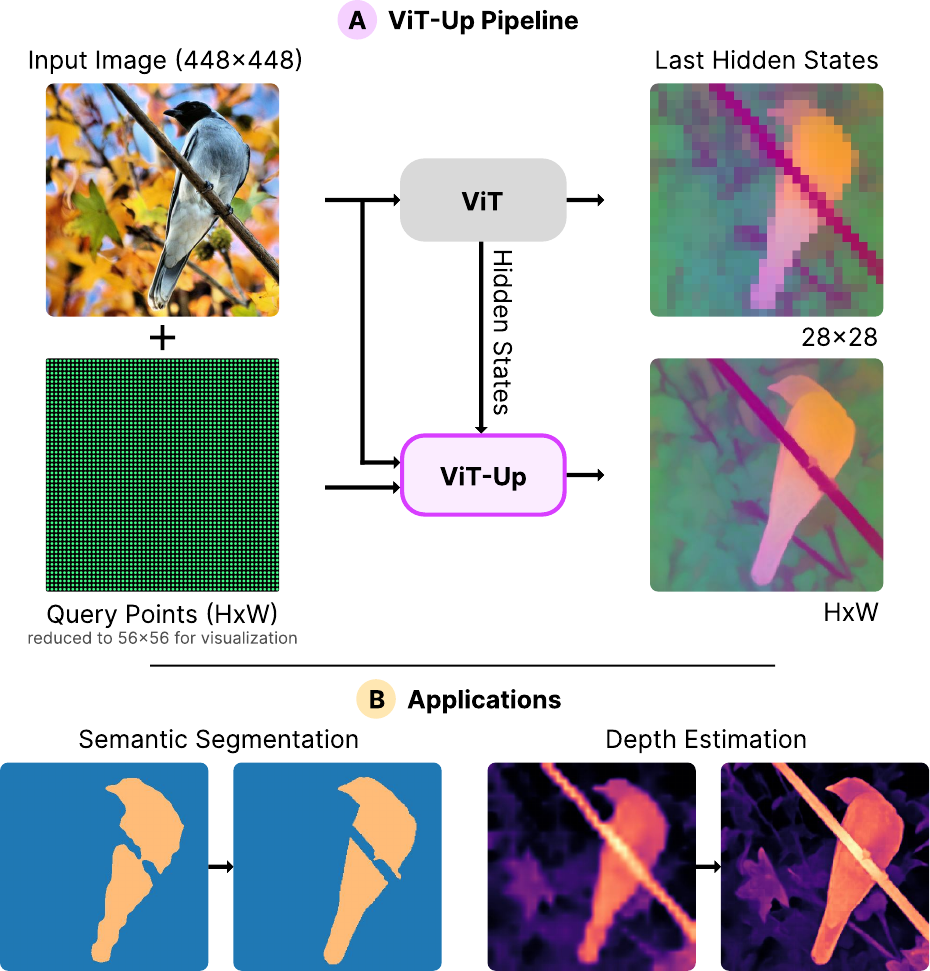}
    \caption{
        \textbf{(A)} Overview of ViT-Up. ViT-Up treats feature upsampling as coordinate-conditioned feature prediction: given an input image and an arbitrary continuous query coordinate, it predicts the corresponding ViT feature from low-resolution backbone hidden states, e.g., the $28{\times}28$ patch-token grid produced by DINOv3-S+ for a $448{\times}448$ input. This enables feature upsampling by independently evaluating the implicit decoder over a dense query grid, yielding high-resolution feature maps at arbitrary output resolutions $H{\times}W$. \textbf{(B)} The resulting dense feature maps can be used for downstream dense prediction tasks such as semantic segmentation and depth estimation.
    }
    \label{fig:hero}
\end{figure}

Vision Transformers (ViTs)~\cite{21_vit} have become the dominant architecture for visual representation learning, providing strong and broadly reusable semantic features~\cite{21_dino, 24_dinov2, 25_dinov3}. By representing images as sequences of patch tokens, ViTs model long-range visual interactions through global self-attention, whose cost scales quadratically with the number of tokens. As a result, foundation vision encoders are commonly evaluated at image resolutions that yield only coarse patch-token grids, for example $14{\times}14$ or $28{\times}28$ tokens. While these low-resolution feature maps provide powerful image representations, they introduce a critical bottleneck for dense prediction tasks, such as semantic segmentation~\cite{16_cityscapes, 17_ade20k} and monocular depth estimation~\cite{24_depth_anything_v2}, which require precise, pixel-level spatial reasoning.

A straightforward remedy is to evaluate the backbone at a higher input resolution, regardless of the computational cost. However, this can move the foundation backbone out-of-distribution and degrade feature quality~\cite{25_jafar}. 
Alternatively, dense prediction systems often use specialized decoders~\cite{15_fcn, 21_dpt, 23_maskdino}. However, these decoders require task-specific training and additional computation, undermining one of the main advantages of foundation models: fast and efficient adaptation to downstream tasks.
Consequently, recent work has introduced task-agnostic feature upsamplers, such as JAFAR~\cite{25_jafar}, AnyUp~\cite{26_anyup}, UpLiFT~\cite{26_uplift}, and NAF~\cite{25_naf}, to bridge the gap between coarse backbone features and dense prediction.

A prevalent strategy is to use high-resolution image guidance. A separate, lightweight image encoder supplies the spatial cues for densifying coarse backbone features. Although the resulting maps can appear sharp, our comparisons reveal feature leakage between visually similar but semantically distinct regions. Disentangling a compressed patch-token grid without mixing adjacent concepts requires the high-resolution guidance signal to possess semantic understanding comparable to the foundation backbone itself. The relatively shallow image encoders of prior work lack this semantic capacity, making image-guided upsampling highly susceptible to feature leakage.

ViT-Up instead exploits the hierarchy of representations already computed by the target backbone. Shallow ViT layers retain spatial and structural detail, while deeper layers organize increasingly abstract semantic content. Together, these states provide the information needed to reconstruct dense features in the backbone's own representation space.

Building on this observation, we introduce ViT-Up, a coordinate-conditioned implicit feature decoder for ViTs (Fig.~\ref{fig:hero}). ViT-Up predicts features at arbitrary continuous image coordinates by progressively refining queries with intermediate ViT hidden states, preserving alignment with the backbone feature space while producing dense outputs beyond the native token grid.

Training proceeds in two stages. In Stage~1, multi-scale backbone teacher features supervise a student that receives a downscaled, padded view of the same image and is queried densely over the visible region, encouraging scale-consistent recovery of spatial detail. In Stage~2, the student predicts spatial offsets to query locally adjusted targets from a frozen Stage~1 ViT-Up teacher, sharpening boundaries and reducing inherited blur and feature leakage. A weak unshifted anchor preserves local detail and spatial structure, while offset regularization discourages large displacements. The decoder is trained once for each target backbone checkpoint and reused across downstream tasks.

We demonstrate that this approach significantly outperforms image-guided methods on standard dense linear probing tasks. On DINOv3-S+, the Stage~2 model achieves $66.42$ mIoU on Cityscapes~\cite{16_cityscapes} and $63.35\%$ $\delta_1$ on COCO depth~\cite{14_coco, 24_depth_anything_v2}, exceeding the strongest evaluated baselines by $3.08$ mIoU and $1.07$ percentage points, respectively (Table~\ref{tab:probing}).

Beyond dense prediction, we evaluate semantic correspondence, a critical measure of feature faithfulness that directly exposes structural weaknesses such as feature blur, fragmentation, and drift~\cite{24_probe3d, 25_dinov3}. On SPair-71k~\cite{19_spair}, Stage~2 reaches $56.39$ PCK@0.10, exceeding the strongest evaluated baseline, bilinear interpolation, by $5.12$ points (Table~\ref{tab:correspondence}). ViT-Up also scales with backbone capacity: with DINOv3-B, the Stage~2 model reaches $70.79$ mIoU on Cityscapes and $58.80$ PCK@0.10 on SPair-71k, widening the margins over the strongest evaluated baselines to $3.47$ mIoU and $8.79$ PCK@0.10 points, respectively (Table~\ref{tab:dinov3_b_results}).

Our contributions are three-fold:
\begin{itemize}
\item We introduce ViT-Up, a task-agnostic implicit feature upsampling framework that predicts backbone-aligned dense feature maps at arbitrary continuous image coordinates. By replacing external image guidance with layer-wise query construction from intermediate ViT representations, ViT-Up reduces common upsampling artifacts such as leakage, fragmentation, and blur.

\item We propose a two-stage training strategy combining multi-scale feature supervision in Stage~1 with offset-guided self-distillation in Stage~2 to recover fine spatial detail and reduce inherited blur and feature leakage while preserving alignment with the backbone feature space.

\item We demonstrate that ViT-Up consistently outperforms prior state-of-the-art feature upsamplers in dense linear probing and semantic correspondence. Moreover, ViT-Up scales effectively with backbone capacity by leveraging richer intermediate representations, unlike image-guided upsamplers whose external guidance signal is largely decoupled from the ViT backbone.
\end{itemize}

Code, pretrained models, and evaluation scripts are available at \url{https://github.com/krispinwandel/vit-up}.

\section{Related Work}

\subsection{Task-Dependent Feature Upsamplers for Dense Prediction}

Dense prediction tasks such as semantic segmentation and monocular depth estimation require spatially detailed feature maps for producing pixel-level predictions. This holds for both CNN-based~\cite{15_fcn,15_unet,17_pspnet} and transformer-based systems~\cite{21_dpt,21_segformer,22_mask2former,23_maskdino}. Since modern backbones often produce features at a lower spatial resolution than the desired output, dense prediction architectures require mechanisms for transforming coarse features into spatially dense representations. For CNNs and hierarchical vision transformers, Feature Pyramid Network (FPN)-style decoders~\cite{17_fpn,18_upernet,21_swin} exploit the native feature hierarchy by upsampling coarse features, typically with bilinear interpolation or learned transposed convolutions, and fusing them with finer backbone features at matching resolutions. Plain ViT backbones, however, do not provide such a spatial hierarchy. Dense prediction systems using these backbones therefore require additional decoder mechanisms to recover spatial resolution: DPT~\cite{21_dpt} reassembles intermediate ViT hidden states into multi-scale decoder features, while ViTDet~\cite{22_vitdet} builds an FPN-like pyramid from a single-scale ViT feature map. In both cases, resolution changes still rely on these standard operations. Dynamic upsampling aims to improve on these standard upsampling operators by predicting input-dependent reassembly weights or sampling locations for local feature aggregation.

CARAFE~\cite{19_carafe,22_carafepp} replaces fixed bilinear interpolation with adaptive feature reassembly, using a shared kernel-prediction module to generate spatially varying kernels conditioned on the local low-resolution feature content. DySample~\cite{23_dysample} replaces dynamic kernels with predicted sampling offsets for grid-based bilinear feature interpolation. Other methods additionally use high-resolution encoder information. FADE~\cite{25_fade} upsamples a low-resolution decoder feature map using kernels generated from both the decoder feature and the corresponding high-resolution encoder skip feature. SAPA~\cite{22_sapa} computes similarities between each high-resolution encoder feature point and a local neighborhood of low-resolution decoder features, and uses these similarities as upsampling weights.

While dynamic upsampling operators can improve dense prediction performance, they are still learned inside task-specific encoder--decoder or feature-pyramid architectures, where upsampling is optimized as one component of a downstream dense prediction model.

This raises the question we target in foundation-model feature upsampling: can feature upsampling be learned once in a task-agnostic way? If a pretrained vision backbone already provides strong semantic features, then a reusable upsampler could densify these features before downstream training. This would make dense adaptation more efficient: instead of training a full high-resolution encoder--decoder model for every task, one could train or finetune lightweight downstream heads on dense foundation-model features with less computation and less task-specific data.

\subsection{Image Super-Resolution and Local Implicit Functions}

A natural source of inspiration is image super-resolution. If low-resolution RGB images can be upsampled to arbitrary resolutions, can low-resolution latent feature maps be upsampled in a similar task-agnostic manner? Local implicit functions~\cite{21_liif} provide an elegant formulation for this question by replacing fixed-grid prediction~\cite{17_edsr} with continuous coordinate-based decoding, enabling the same representation to be sampled at arbitrary locations and densities.

LIIF~\cite{21_liif} encodes the low-resolution input image into a low-resolution grid of latent features, and then predicts the RGB value at arbitrary image coordinates from nearby latent features and their relative offsets to the queried coordinate. The success of this formulation relies on a locality assumption: nearby latent features, and in particular the closest latent cell, contain the information needed to reconstruct the queried high-resolution RGB value.

This assumption is reasonable for image super-resolution because the target signal is a three-channel RGB value and the encoder is trained specifically for local photometric reconstruction. Even though each latent cell supports many sub-pixel details, it only needs to provide enough information for the decoder to predict local color values. This assumption may break in feature upsampling because the target changes from the three-channel RGB domain to a high-dimensional semantic feature space. A low-resolution patch token creates an information bottleneck: it may not have enough capacity to preserve the high-dimensional semantics of all fine-grained details inside its corresponding image patch. As a result, semantic information for high-frequency image detail may be suppressed or mixed before the upsampler is even applied. Thus, the nearest patch token remains a useful local anchor, but it is not sufficient as the sole source of sub-token semantics.

\subsection{Implicit Feature Upsampling}

The optimization-based variant of FeatUp~\cite{24_featup} adapts coordinate-based decoding to vision foundation model features by fitting an implicit neural feature field~\cite{20_nerf, 20_siren} to each input image. It achieves this by iteratively optimizing an MLP through multi-view consistency, leveraging extensive augmentations of the input image. While this approach produces high-quality dense features because the representation is adapted directly to the specific input, the optimization must be repeated for every image. This makes the process prohibitively slow for routine dense evaluation, large-scale probing, or deployment.

LoftUp~\cite{25_loftup} pursues an end-to-end learnable alternative to this per-image optimization. It improves upon local implicit functions~\cite{21_liif} by discarding the restrictive locality assumption where queries rely solely on nearby latent codes. Instead, LoftUp enables high-resolution queries to access the full set of low-resolution feature tokens via cross-attention. Concretely, it adds sinusoidal positional encodings to both the RGB values and the low-resolution feature tokens. The position-enhanced RGB values are projected into the latent feature space using a single $3{\times}3$ convolution, and multiple cross-attention blocks derive the dense features. This global access decouples the query's spatial position from its information source, eliminating the query-to-nearest-cell bottleneck. However, this replaces the locality problem with a semantic retrieval problem. Queries initialized from RGB values and coordinates through a shallow convolution must retrieve semantically appropriate features from the backbone's final hidden state. The shallow projection may not bridge this semantic gap, causing early attention blocks to retrieve incorrect tokens and introduce feature leakage. ViT-Up develops its queries through intermediate backbone states before accessing the final representation. Replacing these intermediate states with repeated access to the final state substantially reduces performance in our ablation (Sec.~IV.L, Table~\ref{tab:ablation}), supporting the value of progressive refinement.

\subsection{Guided Feature Upsampling}

Guided upsampling methods use a high-resolution signal to guide the densification of a lower-resolution target. Classical joint bilateral upsampling~\cite{07_jbu}, guided image filtering~\cite{13_guided_filter}, and fast guided filtering~\cite{16_fast_guided_filter} propagate low-resolution signals while respecting edges and structures in a high-resolution guidance image. FeatUp~\cite{24_featup} was among the first methods to apply this idea directly to vision foundation model features. Its feed-forward variant uses joint bilateral upsampling with the input image as guidance. FeatSharp~\cite{25_featsharp} improves the sharpness of FeatUp's JBU features by combining them with features extracted from a mosaic of higher-resolution image tiles, but this increases feature-extraction cost. Subsequent feature upsamplers demonstrate that sharp dense features can be recovered without repeatedly evaluating the vision backbone on high-resolution tiles.

JAFAR~\cite{25_jafar}, AnyUp~\cite{26_anyup}, and NAF~\cite{25_naf} use a single-stage cross-attention mechanism for image-guided feature upsampling. JAFAR constructs high-resolution image-derived queries and semantically enriched low-resolution keys. AnyUp follows a similar architecture, but introduces a feature-agnostic projection layer that maps low-resolution features of arbitrary dimensionality into the query/key feature space, enabling zero-shot upsampling across different backbones and layers. NAF also targets zero-shot upsampling, but avoids feature-specific key construction by deriving the attention keys from the image encoder alone.

Recursive methods densify features progressively across scales. LiFT~\cite{25_lift} repeatedly upsamples ViT features by fusing the current feature grid with CNN image features extracted at the corresponding image scale. Since each stage operates on the output of the previous one, recursive upsampling can degrade features across scales. UpLiFT~\cite{26_uplift} mitigates this with local attenders that provide image-guided aggregation at each step.

These methods improve spatial sharpness because the guidance signal provides high-resolution boundary and texture cues. However, the semantic quality of the upsampled features remains limited by the semantic understanding of the guidance encoder. If visually similar regions are not sufficiently distinguished by the image-derived guidance signal, the upsampler may aggregate tokens from regions that look alike but correspond to different semantic entities. Moreover, most guided upsamplers operate by reassembling low-resolution token features. Even when the resulting feature maps appear spatially sharper, the features at each high-resolution location are still formed from mixtures of the low-resolution token features, which can make the underlying semantic representation more diffuse.

ViT-Up addresses these limitations by formulating feature upsampling as coordinate-conditioned implicit feature decoding rather than guidance-based token reassembly. Given an arbitrary continuous image coordinate, ViT-Up constructs the corresponding feature from the hierarchy of intermediate ViT hidden states. This keeps the prediction process tied to the backbone feature space while producing dense features beyond the native token grid.

\section{Method}
\label{sec:method}

\begin{figure*}[t]
    \centering
    \includegraphics[width=\textwidth]{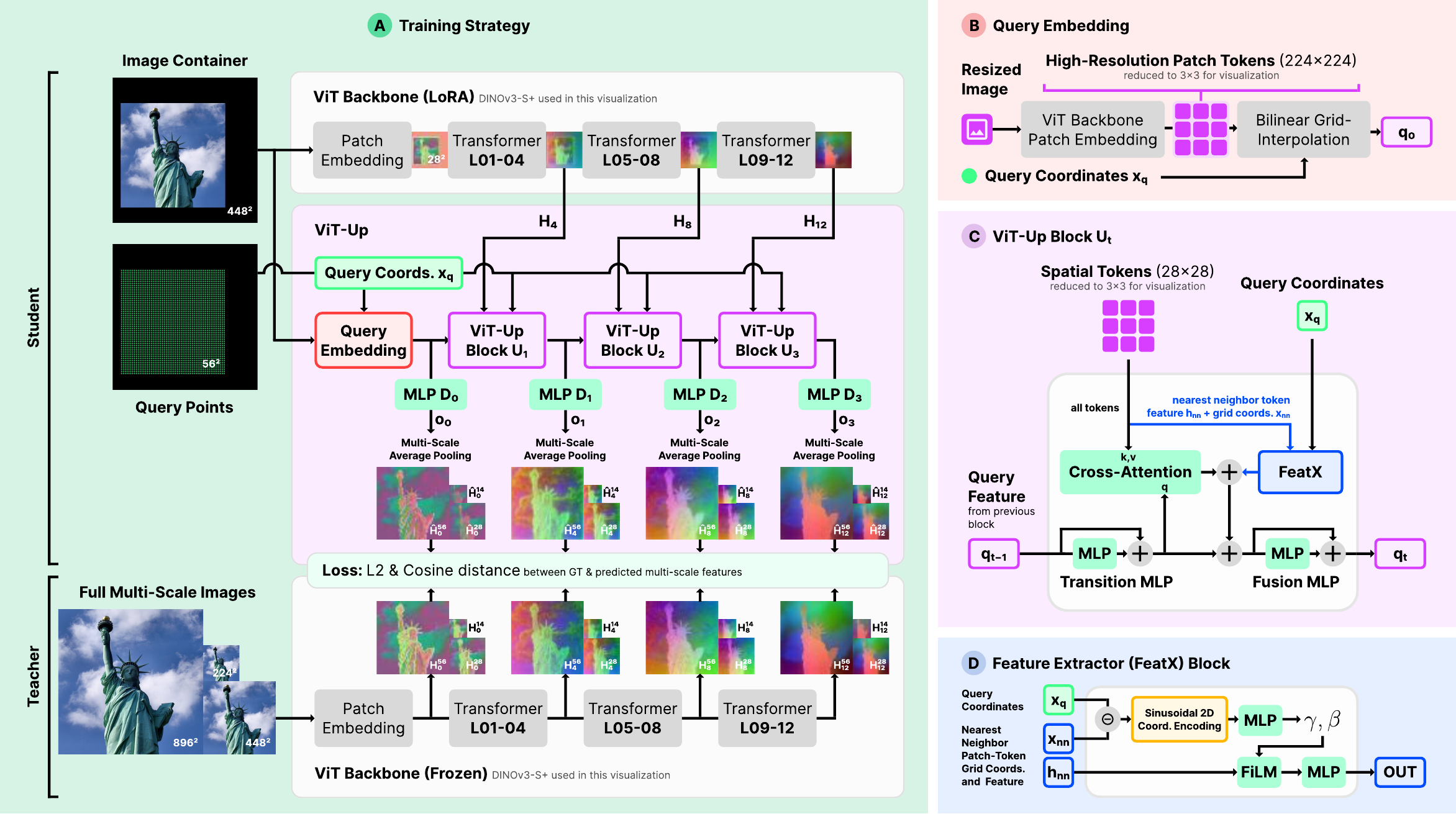}
    \caption{
        Overview of \methodname{}.
        \textbf{(A)} Training strategy.
        A frozen teacher ViT is evaluated on the same image at resolutions $224{\times}224$, $448{\times}448$, and $896{\times}896$, providing feature targets at token-grid resolutions $14{\times}14$, $28{\times}28$, and $56{\times}56$.
        The student receives a downscaled version of the image pasted into a fixed-size image container, together with query coordinates sampled over the visible image region.
        A LoRA-adapted ViT backbone provides low-resolution intermediate hidden states to \methodname{}.
        For each sampled continuous coordinate, \methodname{} first constructs an initial query embedding and then refines it through $T$ ViT-Up blocks that consume the corresponding backbone hidden states.
        For visualization, the figure shows three ViT-Up blocks consuming hidden states from layers $4$, $8$, and $12$.
        In our DINOv3-S+/B main configurations, we use six ViT-Up blocks consuming hidden states from layers $\{2,4,6,8,10,12\}$.
        The decoded query features are supervised at the finest teacher resolution and, after average pooling, at the corresponding coarser teacher token-grid resolutions using feature reconstruction losses.
        \textbf{(B)} Query embedding.
        The input image is processed with the ViT patch-embedding layer to obtain a $224{\times}224$ patch-token grid, which is bilinearly interpolated at the query coordinate $x_q$ to initialize the query representation $q_0$.
        \textbf{(C)} ViT-Up block.
        Each block refines the previous query representation $q_{t-1}$ by fusing token-level context from cross-attention with sub-token detail extracted from the nearest patch token using FeatX.
        \textbf{(D)} FeatX block.
        FeatX encodes the relative offset between the query coordinate and its nearest patch-token center, predicts FiLM parameters, and modulates the nearest-token feature to extract query-specific sub-token information.
    }
    \label{fig:model_arch}
\end{figure*}

\subsection{Problem Formulation}

A pretrained ViT with $L$ transformer layers, patch size $p$, and feature dimension $C$ maps an image of size $(H,W)$ to spatial hidden states $H_l\in\mathbb{R}^{h\times w\times C}$, where $h=H/p$, $w=W/p$, and $l=0,\ldots,L$ includes the embedding layer. Our goal is to predict denser final-layer features $H_L^{\mathrm{up}}\in\mathbb{R}^{h^*\times w^*\times C}$, with $h^*,w^*\gg h,w$, while preserving the backbone's feature-space structure.

\subsection{Architecture}
\label{sec:method_architecture}

ViT-Up predicts a feature at any continuous image coordinate $x_q\in[0,1]^2$, normalized to image boundaries. Evaluating it on a dense query grid yields $H_L^{\mathrm{up}}$ (Fig.~\ref{fig:hero}A). Following the backbone hierarchy, it initializes a query embedding $q_0\in\mathbb{R}^C$ and refines it through $T$ blocks $U_t$, each conditioned on a selected hidden state $H_{l[t]}$:
\begin{equation}
\begin{aligned}
q_0 &= E(x_q,I),\\
q_t &= U_t(q_{t-1},x_q,H_{l[t]}), &&t=1,\ldots,T,\\
o_t^q &= D_t(q_t), &&t=0,\ldots,T.
\end{aligned}
\label{eq:query_refinement}
\end{equation}
Here $I$ is the input image, $E$ the query embedding module, and $D_t$ a LayerNorm followed by a linear projection into the backbone feature space. Only the final output $o_T^q$ is needed at inference. Intermediate outputs support training. Our main configuration uses $T=6$ and $l[t]=2t$, selecting layers $\{2,4,6,8,10,12\}$. Fig.~\ref{fig:model_arch}A illustrates three blocks at layers $\{4,8,12\}$ for clarity.

\paragraph{Query Embedding (Fig.~\ref{fig:model_arch}B)}
We reuse the backbone's patch-embedding convolution, including LoRA when enabled. Applying this single convolution, with kernel size and stride $p$, to the image resized to side $224p$ yields a cached $224\times224$ grid. Bilinear sampling at $x_q$ gives $q_0$.

This image signal initializes queries, while attention keys derive from coarse backbone states rather than shallow image guidance (Sec.~III.B.b). A learned image-independent vector performs comparably (Sec.~IV.L, Table~\ref{tab:ablation}). We retain the shared patch embedding because it adds no parameters and initializes queries in the backbone's representation space.

\paragraph{ViT-Up Block (Fig.~\ref{fig:model_arch}C)}
A transition MLP aligns the incoming query with the current backbone layer, accounting for skipped layers. Cross-attention then supplies token-level context, while FeatX extracts local sub-token detail. Their outputs are fused by a residual MLP:
\begin{equation}
\begin{aligned}
x &= q_{t-1}+\mathrm{MLP}_{\mathrm{transition}}(\mathrm{LN}(q_{t-1})),\\
x^{\mathrm{fused}} &= x+x^{\mathrm{attn}}+x^{\mathrm{sub\text{-}token}},\\
q_t &= x^{\mathrm{fused}}+\mathrm{MLP}_{\mathrm{fusion}}(\mathrm{LN}(x^{\mathrm{fused}})).
\end{aligned}
\end{equation}
For cross-window multi-head attention, we normalize the query and backbone tokens as $\tilde{x}=\mathrm{LN}_Q(x)$ and $\tilde{H}_{l[t]}=\mathrm{LN}_{KV}(H_{l[t]})$, and compute
\begin{equation}
x^{\mathrm{attn}}=W_O\,\mathrm{CW\text{-}MHA}
\left(R_qW_Q\tilde{x},\,R_{\mathbf{X}}W_K\tilde{H}_{l[t]},\,W_V\tilde{H}_{l[t]}\right).
\end{equation}
Continuous two-dimensional RoPE~\cite{24_rope_vit} evaluates the query rotation $R_q$ at $x_q$ and key rotations $R_{\mathbf{X}}$ at patch-token centers $\mathbf{X}$. We implement attention with NATTEN~\cite{23_natten,25_natten_general} or full attention, covering the full $28^2$ backbone grid in the main experiments. Dimensions and attention settings are specified in the supplement.

\paragraph{FeatX (Fig.~\ref{fig:model_arch}D)}
Attention can smooth high-frequency detail, especially in shallow layers. FeatX complements it by decoding the nearest patch token at the query's relative position. Let $k_{\mathrm{nn}}$ index the nearest token center, with coordinate $x_{\mathrm{nn}}=\mathbf{X}[k_{\mathrm{nn}}]$ and feature $h_{\mathrm{nn}}=H_{l[t]}[k_{\mathrm{nn}}]$. We express $\Delta x=(w,h)\odot(x_q-x_{\mathrm{nn}})$ in patch-grid units, equivalent to dividing pixel-coordinate offsets by $p$. Concatenating its two coordinates with $64$ sinusoidal features ($16$ frequency bands per coordinate) gives $p_{\Delta x}=E_{\mathrm{pos}}(\Delta x)\in\mathbb{R}^{66}$~\cite{20_nerf}. Position-conditioned FiLM~\cite{18_film} then modulates the token before decoding:
\begin{equation}
\begin{aligned}
(\gamma,\beta) &= \mathrm{MLP}_{\mathrm{FiLM}}(p_{\Delta x}),\\
\tilde{h}_{\mathrm{nn}} &= (1+\gamma)\odot\mathrm{LN}(h_{\mathrm{nn}})+\beta,\\
x^{\mathrm{sub\text{-}token}} &= \mathrm{MLP}_{\mathrm{sub\text{-}token}}(\tilde{h}_{\mathrm{nn}}).
\end{aligned}
\end{equation}

\paragraph{Backbone Adaptation (Fig.~\ref{fig:model_arch}A)}
We adapt the patch embedding and the attention query, key, value, and output projections with LoRA~\cite{21_lora}, keeping the original backbone weights frozen. For a projection $W$, the update is $W_{\mathrm{adapted}}=W+(\alpha/r)BA$, where $BA$ has rank at most $r$. We use $r=16$, $\alpha=32$, and adapter dropout $0.05$.

\subsection{Training}
\label{sec:training}

\paragraph{Stage 1: Multi-scale Feature Supervision (Fig.~\ref{fig:model_arch}A)}
Very dense backbone inference is expensive and can degrade feature quality~\cite{25_jafar}. We instead distill features from a frozen teacher evaluated at moderate resolutions~\cite{15_hinton_distillation}. The teacher processes the same image at square resolutions $\mathcal{S}=\{224,448,896\}$, producing hidden states $H_l^n$ on token grids with side lengths $n\in\mathcal{N}=\{14,28,56\}$ for $p=16$.

The student receives the image resized to side length $448s$, with $s\sim\mathcal{U}(0.1,1.0)$, and pasted at a random position that keeps the full image within a black $448\times448$ canvas before background augmentation. We query a regular $56\times56$ grid $\{x_{ij}\}$ over the pasted region using Eq.~\eqref{eq:query_refinement}. At each selected layer, decoded outputs form $\hat{H}_{l[t]}^{56}=\{o_t^{ij}\}_{i,j=1}^{56}$. Average pooling supplies the coarser predictions:
\begin{equation}
\hat{H}_{l[t]}^n=\mathrm{AvgPool}_{56\rightarrow n}(\hat{H}_{l[t]}^{56}),
\qquad n\in\mathcal{N}.
\end{equation}
We supervise all refinement stages $t=0,\ldots,T$, with $l[0]=0$ for the embedding layer, against the corresponding $H_{l[t]}^n$, as in hint-based distillation~\cite{14_fitnets}. Recovering these targets from a downscaled view encourages scale-consistent dense features without high-resolution teacher inference.

For a predicted feature $\hat{\mathbf f}$ and teacher feature $\mathbf f$, we combine target-normalized L2 and cosine losses:
\begin{equation}
\begin{aligned}
\mathcal{L}_{\mathrm{L2}}(\hat{\mathbf f},\mathbf f)
&=\frac{\|\hat{\mathbf f}-\mathbf f\|_2^2}{C\,\sigma(\mathbf f)^2},\\
\mathcal{L}_{\mathrm{cos}}(\hat{\mathbf f},\mathbf f)
&=1-\frac{\hat{\mathbf f}^{\top}\mathbf f}{\|\hat{\mathbf f}\|_2\|\mathbf f\|_2+\epsilon},
\end{aligned}
\end{equation}
where $\sigma(\mathbf f)^2=\operatorname{Var}_c(\mathbf f)+\epsilon$ uses the teacher's channel variance at each spatial location and $\epsilon=10^{-5}$. Let $\mathcal{L}_{\mathrm{feat}}=\lambda_{\mathrm{L2}}\mathcal{L}_{\mathrm{L2}}+\lambda_{\mathrm{cos}}\mathcal{L}_{\mathrm{cos}}$, with both weights set to one. When applied to feature maps, this loss averages over spatial locations. The Stage~1 objective is
\begin{equation}
\mathcal{L}_{\mathrm{stage1}}=
\frac{1}{(T+1)|\mathcal{N}|}
\sum_{t=0}^{T}\sum_{n\in\mathcal{N}}
\mathcal{L}_{\mathrm{feat}}(\hat{H}_{l[t]}^n,H_{l[t]}^n).
\end{equation}

\paragraph{Stage 2: Offset-guided refinement}
Stage~1 can inherit blur and feature leakage from its low-resolution targets. We refine it by self-distillation from a frozen Stage~1 ViT-Up teacher, fixing the student padding scale to $0.5$. The teacher sees the full image, while the student learns to recover its finer features from the downscaled view. However, the student's achievable sharpness at this scale remains limited by that of the frozen teacher's full-image features. Simply reducing the scale further shifts the student inputs out of distribution and degrades performance. Updating the teacher with an exponential moving average instead caused drift toward high-frequency channels and reduced semantic fidelity in our experiments.

We therefore keep the teacher fixed and learn corrective query offsets. Let $\mathbf{s}_L(x_q)$ denote the student's final-layer feature at the canvas location corresponding to image coordinate $x_q$, and $\mathbf{t}_L(x_q)$ the teacher's feature at that coordinate. A small MLP predicts $\boldsymbol{\Delta}(x_q)=g_\phi(\mathbf{s}_L(x_q))$, and the teacher is queried at $x_q+\boldsymbol{\Delta}(x_q)$. Near a blurred boundary, an offset can reach a purer object-interior feature, allowing the student to learn a sharper transition at the original location. The offsets are optimized by differentiating the feature-matching loss through the frozen teacher with respect to its query coordinates.

A weak unshifted anchor preserves details that shifted targets could otherwise omit, and an offset penalty discourages large displacements:
\begin{equation}
\begin{aligned}
\mathcal{L}_{\mathrm{stage2}}=\mathbb{E}_{x_q}\big[
&\lambda_{\mathrm{shift}}\,
\mathcal{L}_{\mathrm{feat}}\big(\mathbf{s}_L(x_q),
\mathbf{t}_L(x_q+\boldsymbol{\Delta}(x_q))\big)\\
+{}&\lambda_{\mathrm{anchor}}\,
\mathcal{L}_{\mathrm{feat}}\big(\mathbf{s}_L(x_q),\mathbf{t}_L(x_q)\big)\\
+{}&\lambda_{\mathrm{offset}}\,\|\boldsymbol{\Delta}(x_q)\|_2^2\big].
\end{aligned}
\label{eq:stage2}
\end{equation}
Both matching terms supervise only final-layer features: earlier layers may already be sharp, so offsets correcting final-layer blur could misalign their targets. We use $\lambda_{\mathrm{shift}}=1$, $\lambda_{\mathrm{anchor}}=0.1$, and $\lambda_{\mathrm{offset}}=10$, with image coordinates normalized to $[0,1]^2$. Shifted queries are clipped to the image domain after computing the offset penalty. The inherited student backbone, including LoRA and the shared query initializer, remains frozen. We update the refinement blocks, final decoder, and offset head, using the offset head only during training. The supplement specifies coordinate mapping and optimization details, and visualizes learned offsets and Stage~2 refinement on DINOv2-S and DINOv3-S+.

\paragraph{Dataset and optimization}
We train on ImageNet-1K~\cite{09_imagenet}. Stage~1 uses one epoch, batch size $24$, initial learning rate $2\times10^{-4}$, and cosine annealing. Stage~2 uses $20{,}000$ steps with batch size $16$, about a quarter epoch. Full settings are provided in the supplement and released code.

\section{Experiments}

\begin{figure*}[t]
\centering
\includegraphics[width=\textwidth]{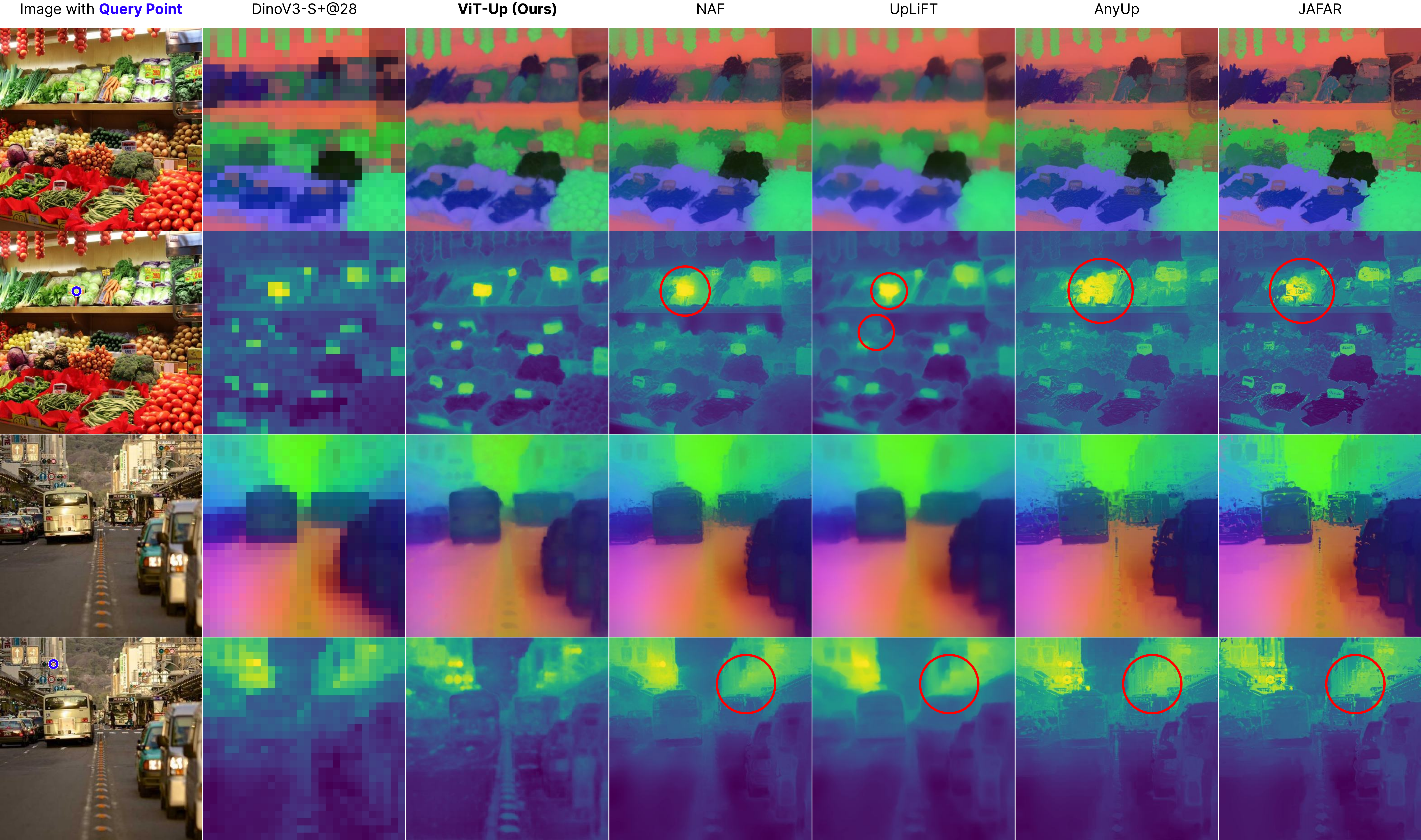}
\caption{
Qualitative comparison of feature upsampling methods on DINOv3-S+.
ViT-Up results use the Stage~1 model.
All methods use a $448{\times}448$ input image. The native backbone produces a $28{\times}28$ feature grid, and upsampled feature maps are shown at $448{\times}448$ output resolution.
We show two examples: a vegetable-store scene in the top two rows and a traffic scene in the bottom two rows.
For each example, the top row visualizes the feature structure using PCA, including the input image, the original $28{\times}28$ backbone feature map, and the upsampled feature maps produced by ViT-Up, NAF~\cite{25_naf}, UpLiFT~\cite{26_uplift}, AnyUp~\cite{26_anyup}, and JAFAR~\cite{25_jafar}.
The bottom row shows the corresponding query-based similarity maps, including the input image with the query point encircled in blue, the similarity map obtained from the low-resolution backbone features, and the similarity maps obtained from the upsampled features of each method.
ViT-Up produces coherent PCA structures and semantically selective similarity maps that remain aligned with the queried region.
In contrast, NAF, AnyUp, and JAFAR can produce visually sharp but fragmented feature maps with leakage into nearby structures, while UpLiFT tends to produce smoother features and weaker similarity responses for small semantic regions.
}
\label{fig:pca_analysis}
\end{figure*}

\subsection{Evaluation Setup}

\paragraph{Backbone}
Our main experiments use DINOv3~\cite{25_dinov3}, as in NAF~\cite{25_naf} and UpLiFT~\cite{26_uplift}. Its cleaner intermediate representations help distinguish upsampling gains from benefits due to suppressing position-dependent artifacts or boundary leakage, as found in DINOv2~\cite{24_dinov2,24_denoising_vit}. However, we also evaluate DINOv2 to assess generalization across backbone families.

\paragraph{Baselines}
We compare with bilinear interpolation, JAFAR~\cite{25_jafar}, AnyUp~\cite{26_anyup}, NAF~\cite{25_naf}, UpLiFT~\cite{26_uplift}, and LoftUp~\cite{25_loftup}. We use official DINOv3-S+ checkpoints for JAFAR and UpLiFT, and train DINOv3-B UpLiFT with its official implementation. AnyUp uses its multi-backbone checkpoint, trained on DINOv2-S, CLIP-B~\cite{21_clip}, SigLIP-B~\cite{23_siglip}, DINOv2-S with registers~\cite{24_registers}, and an ImageNet-supervised ViT-B. NAF uses its released checkpoint, corresponding to the authors' DINOv3-B training configuration. Both support feature-agnostic inference. We encountered reproducibility issues when training LoftUp on DINOv3, so we report its segmentation and correspondence results on DINOv2 only, using its official checkpoint.

To compare with an upsampler of similar capacity, we train a DPT~\cite{21_dpt} dense-feature decoder on frozen DINOv3 with our Stage~1 protocol. It accesses intermediate features and has $18.7$M parameters, compared with ViT-Up's $24.0$M. It bilinearly resizes fused features to $448^2$ before the final convolutions, producing a fixed $448^2$ output. Architecture and training details are in the supplement.

\subsection{Qualitative Analysis}

Fig.~\ref{fig:pca_analysis} compares Stage~1 ViT-Up with image-guided methods through PCA and query-based similarity maps. All methods receive $448^2$ inputs. Dense maps are shown at $448^2$, alongside the native $28^2$ backbone reference. Similarity maps compare the feature nearest the marked query point with every feature in the same map.

The examples distinguish spatial sharpness from semantic selectivity. In the vegetable-store scene, ViT-Up separates neighboring vegetables while matching the queried shelf label to other labels. NAF, AnyUp, and JAFAR instead leak into visually similar green vegetables, despite sharp PCA maps. UpLiFT reduces leakage but blurs several labels. In the traffic scene, ViT-Up selectively matches traffic lights, including small distant instances. The alternatives produce fragmented background responses or suppress small structures.

\subsection{Dense Linear Probing}
\label{sec:dense_linear_probing}
\begin{table*}[t]
\centering
\small
\setlength{\aboverulesep}{0pt}
\setlength{\belowrulesep}{0pt}
\renewcommand{\arraystretch}{1.3}
\setlength{\tabcolsep}{3pt}

\caption[Probing results on DINOv3-S+.]{Dense linear probing on DINOv3-S+ (protocol: Sec.~\ref{sec:dense_linear_probing}). All metrics are $\times100$. Arrows indicate better scores. Green deltas show absolute gains over the best non-ViT-Up baseline (green underline), including RMSE reductions. Stage~2 refines the LoRA model. The top five scores per column are shaded darkest to lightest.}
\label{tab:probing}

\begin{tabular}{l *{10}{C}}
\toprule
& \multicolumn{8}{c}{\textbf{Semantic Segmentation}}
& \multicolumn{2}{c}{\textbf{Depth Estimation}} \\
\cmidrule(lr){2-9}\cmidrule(lr){10-11}
& \multicolumn{2}{c}{COCO}
& \multicolumn{2}{c}{VOC}
& \multicolumn{2}{c}{ADE20K}
& \multicolumn{2}{c}{Cityscapes}
& \multicolumn{2}{c}{COCO} \\
\cmidrule[0.4pt](lr){2-3}\cmidrule[0.4pt](lr){4-5}\cmidrule[0.4pt](lr){6-7}\cmidrule[0.4pt](lr){8-9}\cmidrule[0.4pt](lr){10-11}
\textbf{Method}
& \metric{mIoU$\uparrow$} & \metric{Acc$\uparrow$}
& \metric{mIoU$\uparrow$} & \metric{Acc$\uparrow$}
& \metric{mIoU$\uparrow$} & \metric{Acc$\uparrow$}
& \metric{mIoU$\uparrow$} & \metric{Acc$\uparrow$}
& \metric{$\delta_1\!\uparrow$} & \metric{RMSE$\downarrow$} \\
\midrule

Bilinear & 63.10 & 81.85 & 84.88 & 96.45 & 43.27 & 76.17 & 61.36 & 93.44 & 61.39 & 63.01 \\
DPT
& 61.74 & 81.40
& 84.64 & \bs5{96.46}
& 41.95 & 75.70
& 61.89 & \bb4{94.13}
& 59.83 & 64.36 \\
JAFAR    & 62.50 & 81.50 & 83.88 & 96.16 & 42.48 & 75.81 & 57.78 & 92.47 & 60.98 & 64.35 \\
AnyUp    & 63.03 & 81.83 & 84.54 & 96.34 & 42.77 & 76.02 & 58.96 & 92.93 & 61.56 & 62.54 \\
UpLiFT
& \bs5{63.79} & \bs5{82.28}
& \bs5{85.69} & \bb4{96.72}
& \bb4{44.24} & \bb4{76.71}
& \bs5{63.08} & \bs5{93.94}
& \bs5{61.90} & \bs5{61.32} \\
NAF
& \bb4{63.86} & \bb4{82.33}
& \bb4{85.84} & \bb4{96.72}
& \bs5{44.17} & \bs5{76.69}
& \bb4{63.34} & \bb4{94.13}
& \bb4{62.28} & \bb4{60.69} \\

\arrayrulecolor{rulec}\midrule[0.4pt]\arrayrulecolor{black}

\cellcolor{white}\textbf{ViT-Up} \metric{\footnotesize w/o LoRA}
& \rk3{64.06}{0.20} & \rk3{82.47}{0.14}
& \rk3{86.97}{1.13} & \rk3{97.02}{0.30}
& \rk1{45.10}{0.86} & \rk3{77.00}{0.29}
& \rk2{66.27}{2.93} & \rk2{94.84}{0.71}
& \rk3{62.57}{0.29} & \rk3{59.91}{0.78} \\
\cellcolor{white}\hspace{1em}+ LoRA
& \rk2{64.08}{0.22} & \rk2{82.48}{0.15}
& \rk2{87.45}{1.61} & \rk2{97.13}{0.41}
& \rk3{44.78}{0.54} & \rk2{77.12}{0.41}
& \rk3{65.44}{2.10} & \rk3{94.76}{0.63}
& \rk2{63.12}{0.84} & \rk2{59.19}{1.50} \\
\cellcolor{white}\hspace{1em}+ Stage 2
& \rk1{64.24}{0.38} & \rk1{82.59}{0.26}
& \rk1{87.59}{1.75} & \rk1{97.17}{0.45}
& \rk2{45.00}{0.76} & \rk1{77.32}{0.61}
& \rk1{66.42}{3.08} & \rk1{95.14}{1.01}
& \rk1{63.35}{1.07} & \rk1{58.48}{2.21} \\
\bottomrule
\end{tabular}
\end{table*}

\begin{figure*}[t]
\centering
\includegraphics[width=\textwidth]{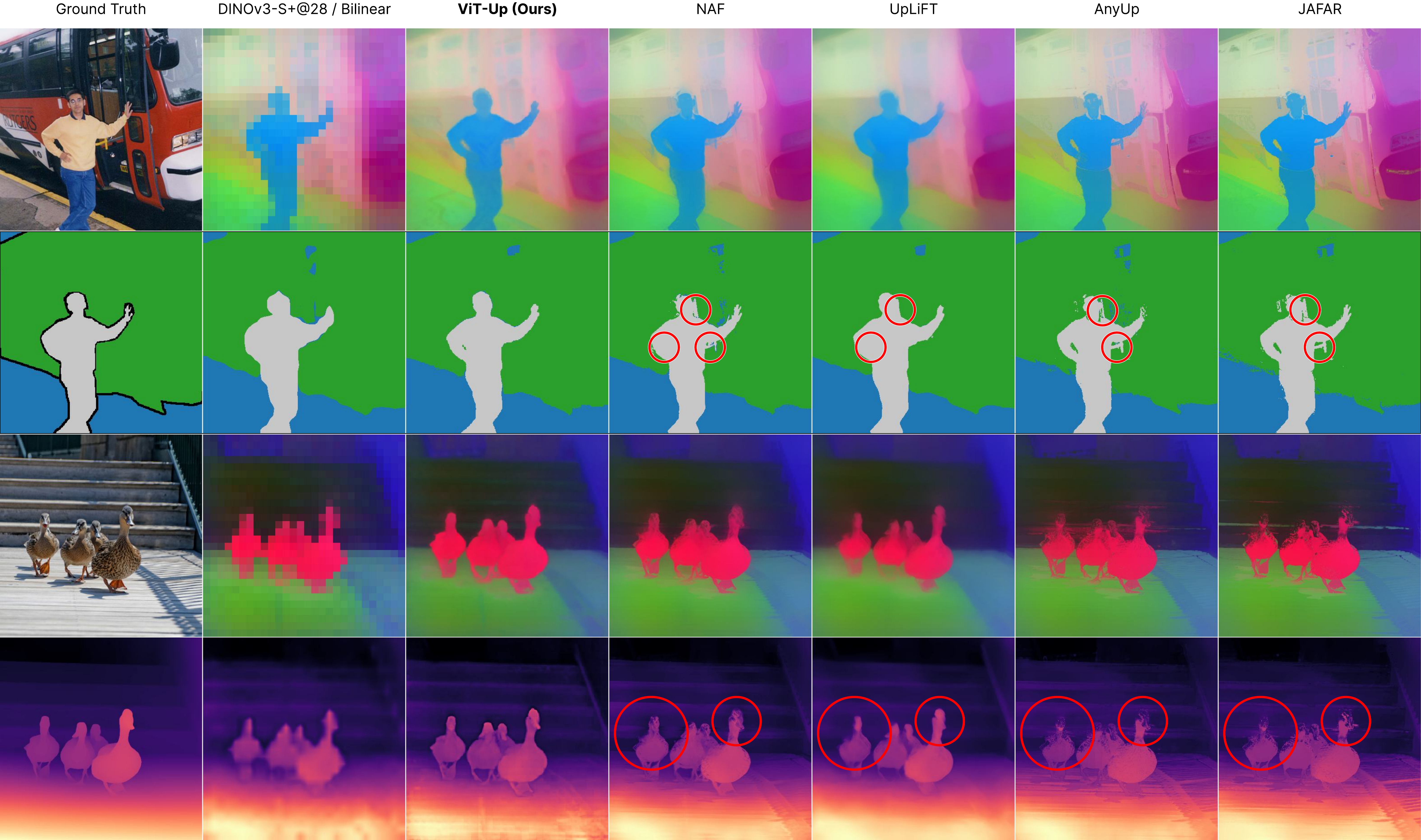}
\caption{
Qualitative dense probing results on DINOv3-S+.
All methods use a $448{\times}448$ input image. The native backbone produces a $28{\times}28$ feature grid, and dense predictions are shown at $448{\times}448$ output resolution.
\textbf{Top two rows:} semantic segmentation example with a person in front of a bus, showing PCA projections of the dense features and the corresponding predicted segmentation masks together with the ground-truth mask (left).
\textbf{Bottom two rows:} monocular depth example with a group of ducks, showing PCA projections of the dense features and the corresponding predicted depth maps together with the pseudo-depth target (left) generated by Depth Anything V2~\cite{24_depth_anything_v2}.
For the bilinear baseline, we visualize PCA on the original $28{\times}28$ backbone feature grid.
NAF~\cite{25_naf}, AnyUp~\cite{26_anyup}, and JAFAR~\cite{25_jafar} produce visually sharp but fragmented feature maps and noisy dense predictions.
UpLiFT~\cite{26_uplift} produces more localized but blurrier features, weakening fine structures such as the person's head region and small distant ducks.
ViT-Up produces more coherent dense features, resulting in cleaner segmentation masks and more consistent depth estimates.
}
\label{fig:probing_analysis}
\end{figure*}

Following JAFAR~\cite{25_jafar}, we use dense linear probing to assess feature quality and sharpness. We freeze each backbone and upsampler and train linear segmentation heads on VOC~\cite{10_voc}, COCO~\cite{14_coco}, ADE20K~\cite{17_ade20k}, and Cityscapes~\cite{16_cityscapes}. For depth, COCO pseudo-targets come from Depth Anything V2~\cite{24_depth_anything,24_depth_anything_v2}. Heads use batch size $4$ and cosine learning-rate decay from $2\times10^{-3}$ for $20$ epochs, or $5$ on COCO. Inputs and prediction grids are $448^2$.

All three ViT-Up variants outperform the baselines across the reported metrics (Table~\ref{tab:probing}). The largest segmentation advantage is on Cityscapes, where Stage~2 exceeds the strongest baseline by $3.08$ mIoU. This gain suggests that the recovered spatial detail helps distinguish the dataset's many small objects and thin structures. Gains on VOC suggest more coherent foreground-object representations, while the depth improvements indicate that the recovered features also retain useful geometric information.

LoRA and Stage~2 have different effects. LoRA improves VOC and depth but lowers ADE20K and Cityscapes mIoU. The frozen-backbone model is already competitive and retains the highest ADE20K score. We hypothesize that ImageNet-1K's emphasis on large foreground objects against a background biases LoRA toward sharper object boundaries, potentially at the expense of small details or semantic fidelity. This could benefit object-centric VOC while hurting Cityscapes, where small objects and fine structures in cluttered scenes are important. Stage~2 improves every reported metric over Stage~1 with LoRA, with the largest segmentation gain again on Cityscapes. Thus, the gains do not depend on backbone adaptation, while refinement provides a more consistent benefit under trained probing.

Fig.~\ref{fig:probing_analysis} connects feature artifacts to prediction errors. Leakage between the person and the yellow bus produces segmentation errors around the upper body in NAF, AnyUp, and JAFAR. UpLiFT instead blurs the head region. In the depth example, fragmented features or excessive smoothing obscure the small distant ducks. ViT-Up better separates the person from the bus and preserves individual duck shapes, illustrating how semantic coherence and spatial detail jointly support dense prediction.

\subsection{Correspondence Estimation}
\label{sec:correspondence_estimation}

\begin{table*}[t]
\centering
\small
\setlength{\aboverulesep}{0pt}
\setlength{\belowrulesep}{0pt}
\renewcommand{\arraystretch}{1.35}
\setlength{\tabcolsep}{5pt}

\caption[Correspondence results on DINOv3-S+.]{DINOv3-S+ correspondence: SPair-71k PCK and NAVI 3D recall (\%). Higher is better. Green/red deltas indicate gains/losses (pp) against the best non-ViT-Up baseline (green underline). Stage~2 refines the LoRA model. The top five scores per column are shaded darkest to lightest.}
\label{tab:correspondence}

\begin{tabular}{l *{6}{C}}
\toprule
& \multicolumn{3}{c}{\textbf{Semantic: SPair-71k, PCK$\uparrow$}}
& \multicolumn{3}{c}{\textbf{Geometric: NAVI, Recall$\uparrow$}} \\
\cmidrule(lr){2-4}\cmidrule(lr){5-7}
\textbf{Method}
& \metric{0.10} & \metric{0.05} & \metric{0.01}
& \metric{5\,cm} & \metric{2\,cm} & \metric{1\,cm} \\
\midrule

Bilinear
& \bb4{51.27} & 33.74 & \bs5{3.83}
& \bs4{80.16} & \bs5{51.18} & \bs5{33.58} \\
DPT
& \bs5{50.86} & \bb4{34.98} & \bb4{6.75}
& 79.53 & \bb4{51.49} & \bb4{34.33} \\
JAFAR
& 36.82 & 18.59 & 1.89
& 79.04 & 47.02 & 26.60 \\
AnyUp
& 37.63 & 19.31 & 1.97
& \bb3{80.31} & 48.78 & 28.37 \\
UpLiFT
& 46.87 & 29.15 & 3.43
& 79.35 & 49.05 & 30.49 \\
NAF
& 48.68 & \bs5{33.96} & 2.89
& 80.03 & 50.29 & 31.62 \\

\arrayrulecolor{rulec}\midrule[0.4pt]\arrayrulecolor{black}

\cellcolor{white}\textbf{ViT-Up} \metric{\footnotesize w/o LoRA}
& \rk3{55.46}{4.19} & \rk3{39.28}{4.30} & \rk3{7.38}{0.63}
& \rk1{81.12}{0.81} & \rk1{53.72}{2.23} & \rk1{36.39}{2.06} \\
\cellcolor{white}\hspace{1em}+ LoRA
& \rk2{55.60}{4.33} & \rk2{39.43}{4.45} & \rk2{7.75}{1.00}
& \rk2{80.67}{0.36} & \rk2{53.21}{1.72} & \rk2{35.93}{1.60} \\
\cellcolor{white}\hspace{1em}+ Stage 2
& \rk1{56.39}{5.12} & \rk1{41.18}{6.20} & \rk1{8.98}{2.23}
& \rl5{80.10}{0.21} & \rk3{52.32}{0.83} & \rk3{35.51}{1.18} \\
\bottomrule
\end{tabular}
\end{table*}

\begin{figure*}[t]
\centering
\includegraphics[width=\textwidth]{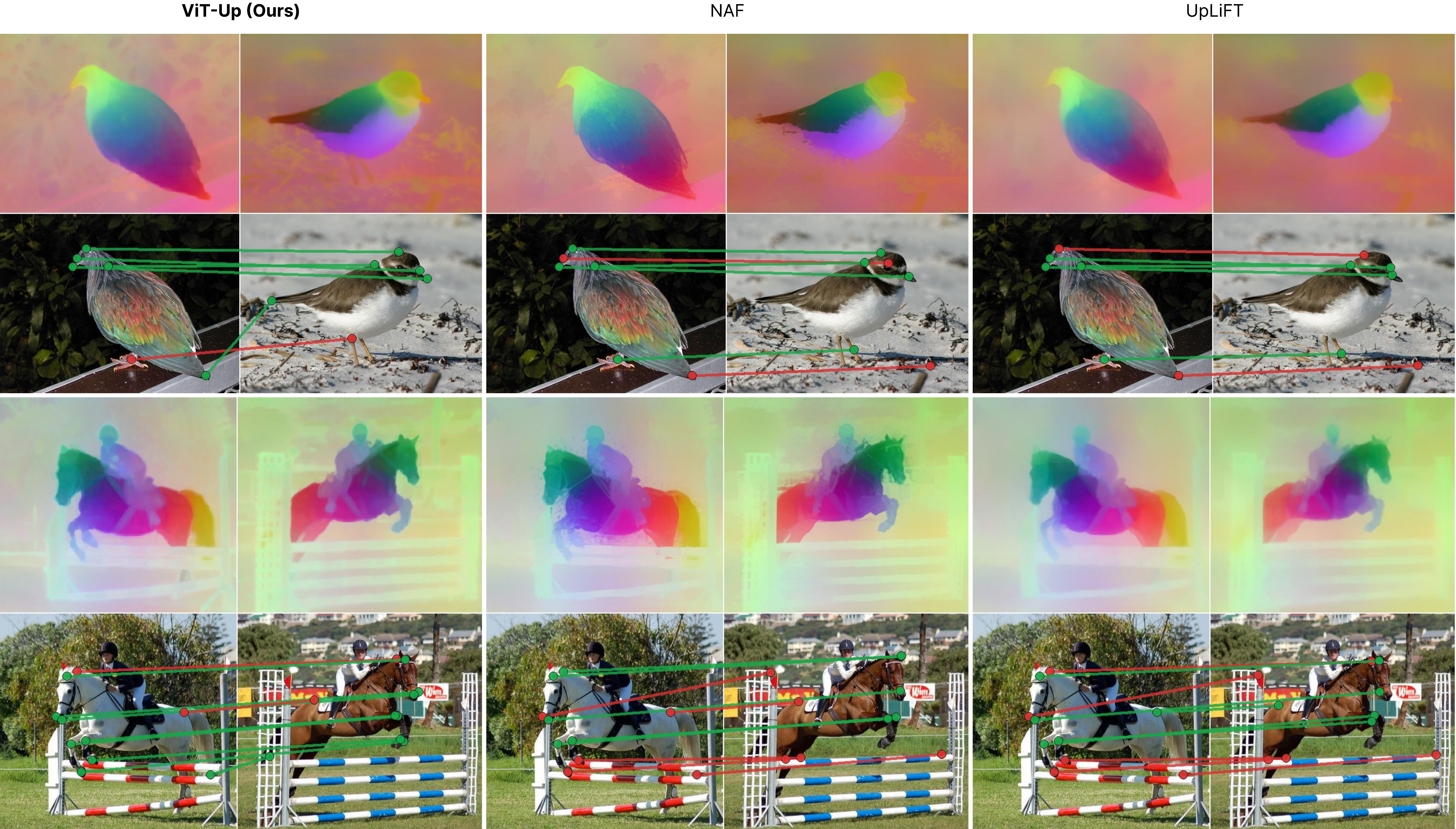}
\caption{
Qualitative semantic correspondence results on SPair-71k.
ViT-Up results use the Stage~1 model.
We show two image pairs, birds in the top two rows and horses in the bottom two rows.
For each pair, the first row shows PCA projections of the dense features, and the second row shows correspondences obtained by matching each source query point to the target location with maximum cosine similarity.
Green lines indicate correct matches under the SPair-71k threshold, while red lines indicate incorrect matches.
ViT-Up produces fine-grained object features and accurate part-level correspondences.
NAF~\cite{25_naf} and UpLiFT~\cite{26_uplift} suffer from feature fragmentation or blurring in small structures such as the bird tail and the partially occluded horse tail, leading to incorrect matches.
}
\label{fig:correspondence}
\end{figure*}

Correspondence tests feature similarity directly, without an adapting task head, and exposes part-level errors that class-level probing can miss. We adapt the DINOv3~\cite{25_dinov3} semantic correspondence protocol on SPair-71k~\cite{19_spair} and the DINOv3/Probe3D~\cite{24_probe3d} geometric protocol on NAVI~\cite{23_navi}.

On SPair-71k, percentage of correct keypoints (PCK) uses normalized distance thresholds $0.10$, $0.05$, and $0.01$. We use $448^2$ inputs and output grids, replacing the original DINOv3 protocol's $1024^2$ inputs. On NAVI, we report 3D correspondence recall within $5$, $2$, and $1$~cm. Inputs are $448^2$ rather than $512^2$. The retained scale factor $0.25$ gives a common $112^2$ feature grid, and we retain $1000$ correspondences. Full matching and coordinate conventions are in the supplement.

All three ViT-Up variants lead on SPair-71k (Table~\ref{tab:correspondence}). Stage~2 improves over the strongest baselines by $5.12$, $6.20$, and $2.23$ PCK points at the three thresholds. At the strictest threshold, it more than doubles bilinear interpolation's score, indicating recovery of precise part-level structure beyond the coarse token grid. LoRA contributes modestly, while Stage~2 gives larger gains, especially at the stricter thresholds.

Fig.~\ref{fig:correspondence} compares Stage~1 ViT-Up, NAF, and UpLiFT. Matches maximize cosine similarity between source and target features. PCA uses a common basis fitted to multi-scale backbone features within object regions obtained from Segment Anything V3 masks~\cite{25_sam3,24_sam2,23_sam}. NAF and UpLiFT mix the bird tail with background features and the partially occluded horse tail with the foreground obstacle. ViT-Up retains more discriminative part features and matches both tails. Its remaining incorrect matches are generally close to the annotated points but outside the PCK@0.10 tolerance. These examples illustrate why visually plausible feature maps need not preserve the local semantic structure required for correspondence.

NAVI shows a different trade-off: Stage~1 without LoRA is the strongest ViT-Up variant, and both adaptation and refinement modestly reduce geometric recall. Stage~2 still exceeds the best baselines at $2$ and $1$~cm but falls $0.21$ points below AnyUp at $5$~cm. LoRA and Stage~2 thus favor semantic part alignment, while the frozen-backbone Stage~1 model better retains cross-view geometric matches.

\subsection{Feature Preservation}

\begin{table}[t]
\centering
\footnotesize
\setlength{\cw}{0.88cm}
\setlength{\aboverulesep}{0pt}
\setlength{\belowrulesep}{0pt}
\renewcommand{\arraystretch}{1.3}
\setlength{\tabcolsep}{2.5pt}

\caption[Feature preservation under frozen-head evaluation.]{Frozen-head evaluation: heads trained on native DINOv3-S+@28 features are applied to upsampled features, with predictions resized to $448^2$. Metrics are $\times100$. Higher is better. Green/red deltas are gains/losses (pp) against the best non-ViT-Up baseline (green underline). Stage~2 refines the LoRA model. The top five scores per column are shaded darkest to lightest.}
\label{tab:feature_preservation}

\begin{tabular}{l *{4}{C} D}
\toprule
& \multicolumn{4}{c}{\textbf{\frozenhead\,Segmentation: mIoU$\uparrow$}}
& \multicolumn{1}{D}{\textbf{\frozenhead\,Depth $\delta_1\!\uparrow$}} \\
\cmidrule(lr){2-5}\cmidrule(lr){6-6}
\textbf{Method}
& \metric{COCO} & \metric{VOC} & \metric{ADE} & \metric{City.}
& \metric{COCO} \\
\midrule

Bilinear
& \bs5{63.10} & \bs5{84.88} & \bs4{43.27} & \bs5{61.36}
& \bs2{61.39} \\
DPT
& 60.07 & 83.21 & 28.95 & 51.56
& 55.79 \\
JAFAR
& 61.97 & 82.35 & 41.79 & 55.15
& 58.71 \\
AnyUp
& 62.88 & 84.47 & 41.89 & 55.18
& \bs3{61.38} \\
UpLiFT
& 63.01 & 83.22 & \bs5{42.80} & 60.01
& \bs5{60.20} \\
NAF
& \bb1{63.78} & \bb4{85.78} & \bb2{43.82} & \bb4{61.41}
& \bb1{62.25} \\

\arrayrulecolor{rulec}\midrule[0.4pt]\arrayrulecolor{black}

\cellcolor{white}\textbf{ViT-Up} \metric{\scriptsize w/o LoRA}
& \rl4{63.61}{0.17} & \rk3{86.68}{0.90} & \rl3{43.80}{0.02} & \rk1{64.73}{3.32}
& \rl0{58.76}{3.49} \\
\cellcolor{white}\hspace{0.8em}+ LoRA
& \rl2{63.71}{0.07} & \rk2{87.02}{1.24} & \rk1{43.98}{0.16} & \rk3{64.01}{2.60}
& \rl4{60.35}{1.90} \\
\cellcolor{white}\hspace{0.8em}+ Stage 2
& \rl3{63.67}{0.11} & \rk1{87.25}{1.47} & \rl0{42.71}{1.11} & \rk2{64.17}{2.76}
& \rl0{59.90}{2.35} \\
\bottomrule
\end{tabular}
\end{table}

We train task heads on native DINOv3-S+@28 features and apply them unchanged to upsampled features (Table~\ref{tab:feature_preservation}). This tests whether increased spatial resolution remains compatible with the feature organization expected by a low-resolution predictor.

All ViT-Up variants lead on VOC and Cityscapes, even exceeding competing methods whose heads were trained on their dense features (Table~\ref{tab:probing}). This supports preservation of class-level semantics. The gains across training stages are not uniform: Stage~1 without LoRA is best on Cityscapes, NAF leads on COCO, and Stage~1 with LoRA leads on ADE20K. Stage~2 improves VOC and Cityscapes over the LoRA model but reduces COCO and ADE20K performance. Refinement therefore adds useful detail without preserving every frozen head's preferred feature statistics.

NAF remains close to its trained-head results. Reassembling final-layer tokens may help retain the feature statistics expected by the original heads. UpLiFT preserves features less well, possibly because deviations accumulate through recursive upsampling.

In frozen-head depth probing, NAF achieves the highest $\delta_1$, while LoRA helps ViT-Up and Stage~2 partly reverses that gain. A depth head trained on coarse tokens expects spatially aggregated features, so more localized features may shift its input statistics. These scores should therefore be interpreted cautiously. They do not by themselves negate the improved class-level and part-level semantics observed in frozen-head segmentation and semantic correspondence, especially since ViT-Up outperforms all baselines when depth heads are trained on its features.

\subsection{Backbone Scaling}

\begin{table}[t]
\centering
\footnotesize
\setlength{\cw}{0.95cm}
\setlength{\aboverulesep}{0pt}
\setlength{\belowrulesep}{0pt}
\renewcommand{\arraystretch}{1.3}
\setlength{\tabcolsep}{2.5pt}

\caption{DINOv3-B segmentation and correspondence (\%, higher is better). ViT-Up uses LoRA. Green deltas show gains (pp) over the best non-ViT-Up baseline (green underline). The top five scores per column are shaded darkest to lightest.}
\label{tab:dinov3_b_results}

\begin{tabular}{l *{5}{C}}
\toprule
& \multicolumn{2}{c}{\textbf{Seg.: mIoU$\uparrow$}}
& \multicolumn{3}{c}{\textbf{SPair-71k: PCK$\uparrow$}} \\
\cmidrule(lr){2-3}\cmidrule(lr){4-6}
\textbf{Method}
& \metric{VOC} & \metric{City.}
& \metric{0.10} & \metric{0.05} & \metric{0.01} \\
\midrule

Bilinear
& \bs5{87.20} & \bs5{64.21} & \bb3{50.01} & \bb3{34.03} & \bb3{3.89} \\
UpLiFT
& \bb3{88.43} & \bb3{67.32} & \bs5{45.37} & \bs4{29.92} & \bs4{3.73} \\
NAF
& \bs4{88.07} & \bs4{66.45} & \bs4{47.19} & \bs5{29.09} & \bs5{3.41} \\

\arrayrulecolor{rulec}\midrule[0.4pt]\arrayrulecolor{black}

\cellcolor{white}\textbf{ViT-Up} \metric{\scriptsize w/ LoRA}
& \rk2{89.18}{0.75} & \rk2{69.81}{2.49}
& \rk2{58.10}{8.09} & \rk2{42.00}{7.97} & \rk2{7.93}{4.04} \\
\cellcolor{white}\hspace{0.8em}+ Stage 2
& \rk1{89.52}{1.09} & \rk1{70.79}{3.47}
& \rk1{58.80}{8.79} & \rk1{43.90}{9.87} & \rk1{9.60}{5.71} \\
\bottomrule
\end{tabular}
\end{table}

For DINOv3-B, we double ViT-Up's internal dimension to match the backbone while retaining the architecture and probing settings. Table~\ref{tab:dinov3_b_results} compares it with bilinear interpolation, NAF, and UpLiFT trained using the official implementation.

Both ViT-Up stages outperform the baselines on every reported metric. UpLiFT improves on both segmentation datasets with the larger backbone and becomes the strongest segmentation baseline on DINOv3-B. Scaling nevertheless increases Stage~2's margin over the strongest baseline from $3.08$ to $3.47$ Cityscapes mIoU and from $5.12$ to $8.79$ SPair-71k PCK@0.10, with wider margins at the stricter correspondence thresholds as well. VOC's margin instead narrows as both ViT-Up and UpLiFT improve. These gains suggest that ViT-Up uses the larger backbone's richer intermediate features to improve fine spatial and part-level discrimination.

A larger backbone alone does not guarantee better correspondence: the bilinear baseline's PCK@0.10 decreases from $51.27$ on DINOv3-S+ to $50.01$ on DINOv3-B. NAF and UpLiFT also decline at this threshold, although UpLiFT improves at the stricter thresholds. The larger backbone may retain more local detail within each patch, making token-level similarities less selective for coarse semantic matching. DINOv3-S+ may instead suppress this detail and emphasize dominant object-level content.

NAF aggregates low-resolution tokens under external image guidance, so it cannot explicitly recover sub-token structure that may be encoded within richer ViT features. Its guidance encoder also gains no capacity when the backbone grows. ViT-Up instead decodes features from intermediate backbone states, enabling it to exploit the additional intra-patch detail.

\subsection{Cross-Backbone Generalization}
\begin{table}[t]
\centering
\footnotesize
\setlength{\cw}{0.95cm}
\setlength{\aboverulesep}{0pt}
\setlength{\belowrulesep}{0pt}
\renewcommand{\arraystretch}{1.3}
\setlength{\tabcolsep}{2.5pt}

\caption{DINOv2-S segmentation and correspondence (\%, higher is better). Green/red deltas show gains/losses (pp) against the best non-ViT-Up baseline (green underline). The top five scores per column are shaded darkest to lightest.}
\label{tab:cross_backbone}

\begin{tabular}{l *{5}{C}}
\toprule
& \multicolumn{2}{c}{\textbf{Seg.: mIoU$\uparrow$}}
& \multicolumn{3}{c}{\textbf{SPair-71k: PCK$\uparrow$}} \\
\cmidrule(lr){2-3}\cmidrule(lr){4-6}
\textbf{Method}
& \metric{VOC} & \metric{City.}
& \metric{0.10} & \metric{0.05} & \metric{0.01} \\
\midrule

Bilinear
& 81.31 & 59.71 & \bb3{53.17} & \bs4{36.05} & \bs4{3.96} \\
AnyUp
& 83.55 & 60.50 & 42.06 & 23.64 & 2.61 \\
LoftUp
& \bb2{84.83} & \bs5{62.36} & 42.81 & 24.89 & 2.89 \\
NAF
& \bs5{84.05} & \bs4{63.78} & \bs5{50.46} & \bs5{31.56} & \bs5{3.86} \\
UpLiFT
& \bs3{84.76} & \bb3{65.30} & \bs4{53.08} & \bb3{36.98} & \bb3{5.66} \\

\arrayrulecolor{rulec}\midrule[0.4pt]\arrayrulecolor{black}

\cellcolor{white}\textbf{ViT-Up} \metric{\scriptsize w/o LoRA}
& \rl4{84.15}{0.68} & \rk2{65.82}{0.52} & \rk1{54.43}{1.26} & \rk2{39.98}{3.00} & \rk2{7.37}{1.71} \\
\cellcolor{white}\hspace{0.8em}+ Stage 2
& \rk1{84.96}{0.13} & \rk1{67.42}{2.12} & \rk2{53.98}{0.81} & \rk1{40.35}{3.37} & \rk1{8.70}{3.04} \\
\bottomrule
\end{tabular}
\end{table}

\label{sec:cross_backbone}

To assess ViT-Up's generalization to other backbone families, we also test on DINOv2-S. Compared with DINOv3, its features exhibit more pronounced position-dependent artifacts and stronger boundary leakage. To reduce the former, Stage~1 uses teacher features from a $50\%$ crop when the sampled padding scale is below $0.5$, and from the full image otherwise. Stage~2 then reduces inherited leakage and blur. Qualitative comparisons of both changes are provided in the supplement.

With Stage~2, ViT-Up leads segmentation on both VOC and Cityscapes and exceeds the best baseline by $3.04$ points at the strictest SPair-71k threshold (Table~\ref{tab:cross_backbone}). Stage~1 remains best at the coarsest threshold, while Stage~2 improves both stricter thresholds. DINOv2's stronger initial leakage may explain why Stage~2 yields a larger VOC gain than on DINOv3-S+. Qualitatively, LoftUp sharpens boundaries but smooths intra-object features, which may reflect its mask-based supervision. UpLiFT retains more internal detail, while NAF and AnyUp show boundary leakage. ViT-Up combines sharper boundaries with discriminative internal structure, supporting both segmentation and fine-grained correspondence.

\subsection{Backbone Resolution}
\label{sec:backbone_resolution}

\begin{table}[t]
\centering
\footnotesize
\setlength{\cw}{0.8cm}
\setlength{\aboverulesep}{0pt}
\setlength{\belowrulesep}{0pt}
\renewcommand{\arraystretch}{1.3}
\setlength{\tabcolsep}{2.5pt}

\caption{Backbone-resolution comparison on DINOv3-S+. Grid denotes the native token grid. ViT-Up decodes $448^2$ features. ViT-Up$^{112^2}$ + Bilinear decodes $112^2$, then resizes logits for segmentation and features for correspondence and timing to $448^2$. Task metrics are in \%. The top five ranks are shaded darkest to lightest, with the best bold. Ties share a rank.}
\label{tab:ablation_backbone_resolution}

\begin{adjustbox}{max width=\columnwidth}
\begin{tabular}{l c c *{5}{C}}
\toprule
& & & \multicolumn{2}{c}{\textbf{Seg.: mIoU$\uparrow$}}
& \multicolumn{3}{c}{\textbf{SPair-71k: PCK$\uparrow$}} \\
\cmidrule(lr){4-5}\cmidrule(lr){6-8}
\textbf{Method} & \textbf{Grid} & \textbf{Time [ms]$\downarrow$}
& \metric{VOC} & \metric{City.}
& \metric{0.10} & \metric{0.05} & \metric{0.01} \\
\midrule

\multirow{3}{*}{Bilinear}
& $28^2$ & \bs1{\textbf{3.7}}
& 84.9 & 61.4 & \bs5{51.3} & \bs5{33.7} & 3.8 \\
& $56^2$ & \bs2{3.9}
& \bs5{87.0} & \bs3{67.4} & \bs4{56.0} & \bs4{40.2} & \bs5{7.4} \\
& $112^2$ & \bs5{15.9}
& 82.7 & 61.9 & 49.8 & 33.1 & 6.3 \\

\arrayrulecolor{rulec}\midrule[0.4pt]\arrayrulecolor{black}

\multirow{2}{*}{\shortstack[l]{\textbf{ViT-Up}$^{112^2}$\\+ Bilinear}}
& $28^2$ & \bs3{9.5}
& \bs3{87.5} & \bs5{65.7} & \bs3{56.3} & \bs3{41.0} & \bs4{8.9} \\
& $56^2$ & \bs4{10.5}
& \bs4{87.2} & \bs2{68.7} & \bs2{56.4} & \bs3{41.0} & \bs2{9.2} \\

\arrayrulecolor{rulec}\midrule[0.4pt]\arrayrulecolor{black}

\multirow{2}{*}{\textbf{ViT-Up}}
& $28^2$ & 46.1
& \bs2{87.6} & \bs4{66.4} & \bs2{56.4} & \bs2{41.2} & \bs3{9.0} \\
& $56^2$ & 59.7
& \bs1{\textbf{87.8}} & \bs1{\textbf{70.6}} & \bs1{\textbf{58.0}} & \bs1{\textbf{42.9}} & \bs1{\textbf{9.9}} \\
\bottomrule
\end{tabular}
\end{adjustbox}
\end{table}

ViT-Up and higher-resolution backbone inference are complementary. Table~\ref{tab:ablation_backbone_resolution} compares bilinear upsampling of DINOv3-S+ features on $28^2$, $56^2$, and $112^2$ backbone grids with ViT-Up on the first two grids. Evaluation follows the main benchmark protocol. ViT-Up either decodes $448^2$ features directly or decodes $112^2$ followed by bilinear resizing to $448^2$. The $56^2$ backbone uses a model trained jointly on $28^2$ and $56^2$ grids.

On both $28^2$ and $56^2$ backbone grids, ViT-Up$^{112^2}$ + Bilinear improves every task metric over bilinear upsampling from the same backbone grid. With the larger grid, Cityscapes improves by $1.3$ mIoU. Direct $448^2$ decoding brings further gains. Even with a $28^2$ backbone, ViT-Up exceeds the $56^2$ bilinear baseline on VOC and all correspondence thresholds. Increasing the backbone grid itself from $56^2$ to $112^2$ instead lowers every task score, possibly due to a mismatch with pretraining resolutions. Thus, ViT-Up can both recover detail from coarse grids and exploit the additional information in denser backbone representations.

At a $56^2$ backbone grid, quadratic attention remains manageable, and backbone inference followed by bilinear upsampling takes $3.9$~ms. The quadratic cost becomes increasingly restrictive as the grid grows, with the $112^2$ backbone baseline taking $15.9$~ms. ViT-Up increases output density independently of the backbone grid: the $112^2$ + Bilinear variant takes $9.5$ and $10.5$~ms on $28^2$ and $56^2$ backbone grids, respectively, while outperforming the $112^2$ backbone baseline on every task metric.

\subsection{Output Feature Resolution}

\begin{table}[t]
\centering
\footnotesize
\setlength{\cw}{0.95cm}
\setlength{\aboverulesep}{0pt}
\setlength{\belowrulesep}{0pt}
\renewcommand{\arraystretch}{1.3}
\setlength{\tabcolsep}{2.5pt}

\caption{Stage~2 output-resolution sweep on DINOv3-S+: frozen-head segmentation mIoU and SPair-71k PCK (\%). Heads trained on raw $28\times28$ backbone features remain fixed. Segmentation logits and both source and target correspondence features are bilinearly resized to $448^2$. Higher is better. Ranks are shaded darkest to lightest, with the best bold. Ties share a rank.}
\label{tab:ablation_output_resolution}

\begin{tabular}{l *{5}{C}}
\toprule
& \multicolumn{2}{c}{\textbf{\frozenhead\,Seg.: mIoU$\uparrow$}}
& \multicolumn{3}{c}{\textbf{SPair-71k: PCK$\uparrow$}} \\
\cmidrule(lr){2-3}\cmidrule(lr){4-6}
\textbf{Decoded grid}
& \metric{VOC} & \metric{City.}
& \metric{0.10} & \metric{0.05} & \metric{0.01} \\
\midrule

$28{\times}28$
& \bs4{84.82} & \bs5{56.66} & \bs5{50.81} & \bs5{33.38} & \bs5{3.75} \\
$56{\times}56$
& \bs3{86.85} & \bs4{61.38} & \bs4{55.06} & \bs4{39.20} & \bs4{6.96} \\
$112{\times}112$
& \bs2{87.25} & \bs3{63.51} & \bs3{56.33} & \bs3{41.01} & \bs3{8.93} \\
$224{\times}224$
& \bs1{\textbf{87.26}} & \bs2{64.03} & \bs1{\textbf{56.49}} & \bs1{\textbf{41.24}} & \bs1{\textbf{9.15}} \\
$448{\times}448$
& \bs2{87.25} & \bs1{\textbf{64.17}} & \bs2{56.39} & \bs2{41.18} & \bs2{8.98} \\
\bottomrule
\end{tabular}
\end{table}

\begin{table}[t]
\centering
\footnotesize
\setlength{\cw}{0.84cm}
\setlength{\aboverulesep}{0pt}
\setlength{\belowrulesep}{0pt}
\renewcommand{\arraystretch}{1.3}
\setlength{\tabcolsep}{2.5pt}

\caption{Stage~2 dense probing at two output resolutions on DINOv3-S+. Inputs and targets remain $448^2$. Segmentation logits and depth-bin scores are bilinearly resized to $448^2$. Metrics are $\times100$. Higher is better. Best scores are bold and shaded darker.}
\label{tab:vitup_resolution_probing}

\begin{tabular}{l *{4}{C} D}
\toprule
& \multicolumn{4}{c}{\textbf{Segmentation: mIoU$\uparrow$}}
& \multicolumn{1}{D}{\textbf{Depth $\delta_1\!\uparrow$}} \\
\cmidrule(lr){2-5}\cmidrule(lr){6-6}
\textbf{Resolution}
& \metric{COCO} & \metric{VOC} & \metric{ADE} & \metric{City.}
& \metric{COCO} \\
\midrule

$112{\times}112$
& \bs3{64.23} & \bs3{87.52} & \bs3{44.91} & \bs3{65.73}
& \bs3{63.29} \\
$448{\times}448$
& \bs1{\textbf{64.24}} & \bs1{\textbf{87.59}} & \bs1{\textbf{45.00}} & \bs1{\textbf{66.42}}
& \bs1{\textbf{63.35}} \\
\bottomrule
\end{tabular}
\end{table}

We study output resolution using the same Stage~2 model, $448^2$ inputs, and a $28^2$ backbone grid. Table~\ref{tab:ablation_output_resolution} evaluates frozen-head segmentation and semantic correspondence, while Table~\ref{tab:vitup_resolution_probing} uses standard dense linear probing with heads trained at each resolution. In both tables, lower-resolution outputs are bilinearly interpolated to $448^2$ for evaluation.

At $28^2$ output, SPair-71k scores are only slightly below the $28^2$ bilinear baseline. Coarse sampling of ViT-Up's more localized features may omit source or target cues retained in spatially aggregated backbone tokens, which interpolation cannot recover. SPair-71k improves with output resolution up to $224^2$ and then saturates. Even at the strictest PCK threshold, the median target bounding box gives an accepted error of $2.9$ pixels on the $448^2$ canvas. Finer localization within this radius leaves PCK unchanged, which may help explain the plateau.

Frozen-head Cityscapes continues to improve from $224^2$ to $448^2$, gaining $0.14$ mIoU. VOC largely plateaus. Its boundary void band~\cite{10_voc} excludes many fine refinements from evaluation. With separately trained heads, all reported segmentation and depth metrics improve from $112^2$ to $448^2$, with the largest segmentation gain on Cityscapes. Denser querying therefore recovers useful detail even at $448^2$.

\subsection{Convergence Analysis}

\begin{figure*}[t]
\centering
\begingroup
\definecolor{lorablue}{HTML}{4F6BED}
\definecolor{frozenorange}{HTML}{CF7B4C}
\definecolor{darktext}{HTML}{111111}
\definecolor{mutedtext}{HTML}{4B5563}
\definecolor{softaxis}{HTML}{B8C1CD}
\definecolor{softgrid}{HTML}{E4E9F0}
\definecolor{panelbg}{HTML}{F8FAFC}
\pgfplotsset{
  convergencestyle/.style={
    width=2.7cm,height=2.5cm,scale only axis,clip mode=individual,
    xmin=0.105,xmax=4.76,xtick={0.125,0.25,0.5,1,2,4},
    xticklabels={$\frac{1}{8}$,$\frac{1}{4}$,$\frac{1}{2}$,$1$,$2$,$4$},
    xlabel={},ylabel={},
    font=\sffamily,
    axis lines=left,
    axis line style={draw=softaxis,line width=0.5pt},
    axis background/.style={fill=panelbg},
    tick align=outside,major tick length=2pt,
    tick style={draw=softaxis,line width=0.5pt},
    tick label style={font=\sffamily\footnotesize,text=mutedtext},
    xticklabel style={yshift=-1pt},
    yticklabel style={xshift=-1pt,
      /pgf/number format/fixed,/pgf/number format/fixed zerofill,
      /pgf/number format/precision=1},
    scaled y ticks=false,
    ymajorgrids=true,grid style={draw=softgrid,line width=0.45pt},
    title style={font=\sffamily\small,text=darktext,yshift=3pt},
  },
  loraline/.style={color=lorablue,solid,line width=1.2pt,
    mark=*,mark size=2.4pt,
    mark options={solid,fill=lorablue,draw=white,line width=0.5pt}},
  frozenline/.style={color=frozenorange,dash pattern=on 3pt off 2.2pt,line width=1.2pt,
    mark=square*,mark size=2.3pt,
    mark options={solid,fill=white,draw=frozenorange,line width=0.9pt}},
}
\def\convpanel#1#2#3#4#5#6#7#8{%
\begin{axis}[xmode=log,log basis x=2,xminorticks=false,
  convergencestyle,at={(#1cm,0cm)},anchor=south west,
  ymin=#4,ymax=#5,ytick={#6}]
\node[anchor=south west,inner sep=0pt,xshift=-0.78cm,yshift=3pt,font=\sffamily\footnotesize,text=darktext]
  at (rel axis cs:0,1) {#2};
\node[anchor=south,inner sep=0pt,yshift=3pt,font=\sffamily\footnotesize,text=darktext]
  at (rel axis cs:0.5,1) {#3};
\addplot[loraline] coordinates {#7};
\addplot[frozenline] coordinates {#8};
\end{axis}}
\def\convheader#1#2#3{%
\node[anchor=south,font=\sffamily\small,text=darktext] at ({(#1+#2)/2},3.0) {#3};}
\def\keylora{\tikz[baseline=-0.6ex]{%
  \draw[color=lorablue,line width=1.2pt] (0,0) -- (0.75,0);
  \filldraw[fill=lorablue,draw=white,line width=0.5pt] (0.375,0) circle[radius=2.4pt];}}
\def\keyfrozen{\tikz[baseline=-0.6ex]{%
  \draw[color=frozenorange,dash pattern=on 3pt off 2.2pt,line width=1.2pt] (0,0) -- (0.75,0);
  \filldraw[fill=white,draw=frozenorange,line width=0.9pt]
    (0.375-0.083,-0.083) rectangle (0.375+0.083,0.083);}}
\def\keyarrow{\tikz[baseline=-0.6ex]{%
  \draw[-{Latex[length=7pt,width=6pt]},color=darktext,line width=0.8pt] (0,0) -- (0.95,0);}}
\resizebox{\textwidth}{!}{%
\begin{tikzpicture}
\convpanel{0}{mIoU}{}{86.2}{87.6}{86.4,86.8,87.2}
  {(0.125,86.8425488472) (0.25,87.2589230537) (0.5,87.4188184738) (1,87.4465823174) (2,87.3972415924) (4,87.4139487743)}
  {(0.125,86.3982796669) (0.25,86.6559922695) (0.5,86.8350803852) (1,86.9691848755) (2,87.0135486126) (4,87.0403051376)}
\convpanel{4.4}{mIoU}{}{63.5}{67.0}{64.0,65.0,66.0}
  {(0.125,63.8866186142) (0.25,64.8776888847) (0.5,65.1490330696) (1,65.4433608055) (2,65.5668973923) (4,65.7560706139)}
  {(0.125,64.9519562721) (0.25,65.5786752701) (0.5,66.0738945007) (1,66.2742078304) (2,66.5058195591) (4,66.6401863098)}
\convpanel{8.8}{PCK}{$\alpha{=}0.10$}{54.3}{55.8}{54.5,55.0,55.5}
  {(0.125,54.5781877592) (0.25,55.3101460419) (0.5,55.5160939573) (1,55.5976218689) (2,55.5668953349) (4,55.4151539467)}
  {(0.125,54.7422193798) (0.25,55.1950119529) (0.5,55.3299722966) (1,55.4617413925) (2,55.5748961257) (4,55.4613565990)}
\convpanel{12.7}{PCK}{$\alpha{=}0.05$}{37.0}{39.8}{37.0,38.0,39.0}
  {(0.125,37.1611385113) (0.25,38.4827040372) (0.5,39.2380067337) (1,39.4320040386) (2,39.3694495939) (4,39.3392936437)}
  {(0.125,37.5413168922) (0.25,38.4864061134) (0.5,39.1534794456) (1,39.2754463878) (2,39.5523356135) (4,39.5487475450)}
\convpanel{16.6}{PCK}{$\alpha{=}0.01$}{5.7}{8.3}{6.0,7.0,8.0}
  {(0.125,5.9363774391) (0.25,6.7921005432) (0.5,7.3545998153) (1,7.7523884287) (2,7.7851366210) (4,7.8972712357)}
  {(0.125,6.0713319477) (0.25,6.5173599983) (0.5,6.9561227074) (1,7.3800073277) (2,7.7624365349) (4,8.0615983386)}
\convheader{0}{2.7}{VOC}
\convheader{4.4}{7.1}{Cityscapes}
\convheader{8.8}{19.3}{SPair-71k}
\node[font=\sffamily\small,text=darktext,anchor=north] at (9.65cm,-0.85cm) {%
  \keylora\hspace{0.5em}With LoRA\hspace{1.8em}%
  \keyfrozen\hspace{0.5em}Without LoRA\hspace{1.8em}%
  {\color{softaxis}\rule[-0.5ex]{0.5pt}{1.9ex}}\hspace{1.8em}%
  Training budget (epochs)\hspace{0.6em}\keyarrow};
\end{tikzpicture}%
}
\endgroup
\caption{Convergence of Stage~1 ViT-Up on DINOv3-S+, with and without rank-16 LoRA. Training budgets range from $\frac{1}{8}$ to $4$ ImageNet-1K epochs. VOC and Cityscapes report trained-probe mIoU. PCK is measured on SPair-71k. All scores are percentages. Each point reports results from the checkpoint with the lowest validation feature loss in a separate run of the indicated duration. The epoch axis is log$_2$-scaled and each panel uses its own zoomed $y$-range.}
\label{fig:lora-convergence}
\end{figure*}

Fig.~\ref{fig:lora-convergence} compares Stage~1 training for $\frac18$, $\frac14$, $\frac12$, $1$, $2$, and $4$ ImageNet-1K epochs, with and without LoRA. Each budget is a separate run from scratch with its own cosine schedule, fixed warmup duration, and checkpoint selection by minimum validation feature-reconstruction loss.

With LoRA, one epoch is within approximately $0.1$ points of the best tested VOC and PCK@0.10/0.05 scores, supporting our default training budget. From one to four epochs, both variants gain approximately $0.3$--$0.4$ mIoU on Cityscapes. Across all tested budgets, the frozen backbone leads on Cityscapes while LoRA leads on VOC, with no consistent winner across datasets. Without LoRA, some correspondence metrics continue improving to match or exceed LoRA at the stricter thresholds after two to four epochs. As discussed in Sec.~\ref{sec:dense_linear_probing}, LoRA may sharpen foreground boundaries at the expense of fine detail or semantic fidelity, helping explain its persistent VOC gain and Cityscapes deficit.

\subsection{Video Object Segmentation}
\label{sec:video_segmentation}

\begin{table}[t]
\centering
\footnotesize
\setlength{\cw}{1.75cm}
\setlength{\aboverulesep}{0pt}
\setlength{\belowrulesep}{0pt}
\renewcommand{\arraystretch}{1.3}
\setlength{\tabcolsep}{1.5pt}

\caption{DAVIS-2017 video object segmentation ($\mathcal{J}\&\mathcal{F}$, \%) with linear probing (mean over $10$ seeds) and k-NN label propagation. Deltas, underlines, and shading follow Table~\ref{tab:cross_backbone}. LILA uses its official DINOv2-S checkpoint.}
\label{tab:video_segmentation}

\begin{tabular}{l *{4}{C}}
\toprule
& \multicolumn{2}{c}{\textbf{DINOv2-S}}
& \multicolumn{2}{c}{\textbf{DINOv3-S+}} \\
\cmidrule(lr){2-3}\cmidrule(lr){4-5}
\textbf{Method}
& \metric{Linear Probing} & \metric{k-NN}
& \metric{Linear Probing} & \metric{k-NN} \\
& \metric{$\mathcal{J}\&\mathcal{F}\!\uparrow$} & \metric{$\mathcal{J}\&\mathcal{F}\!\uparrow$}
& \metric{$\mathcal{J}\&\mathcal{F}\!\uparrow$} & \metric{$\mathcal{J}\&\mathcal{F}\!\uparrow$} \\
\midrule

Bilinear
& \bs5{58.6} & \bs5{62.2} & \bs4{63.7} & \bs3{63.3} \\
LILA
& \bs3{64.5} & \bb2{73.9} & -- & -- \\
NAF
& \bs4{62.6} & \bs4{65.9} & \bs3{63.9} & \bs4{62.8} \\
UpLiFT
& \bb2{65.1} & \bs3{69.9} & \bb2{66.0} & \bb2{67.4} \\

\arrayrulecolor{rulec}\midrule[0.4pt]\arrayrulecolor{black}

\cellcolor{white}\textbf{ViT-Up}
& \rk1{67.3}{2.2} & \rk1{76.0}{2.1} & \rk1{69.5}{3.5} & \rk1{73.1}{5.7} \\
\bottomrule
\end{tabular}
\end{table}

Table~\ref{tab:video_segmentation} evaluates temporal consistency on the $30$ DAVIS-2017~\cite{17_davis} validation videos, following LILA's~\cite{26_lila} linear-probing and local k-NN protocols, which transfer the first frame's object masks to later frames. Rather than combining bilinear and dense outputs as in LILA, we use only dense outputs to isolate each upsampler's contribution; the supplement reports the combined setting. $\mathcal{J}\&\mathcal{F}$ averages the mean region IoU $\mathcal{J}$ and the mean boundary F-measure $\mathcal{F}$. We use the Stage~2 ViT-Up models trained jointly on $28^2$ and $56^2$ grids (Sec.~\ref{sec:backbone_resolution}), or $32^2$ and $64^2$ for DINOv2-S. For LILA, we use the official DINOv2-S checkpoint; we could not reproduce its training on DINOv3-S+, as its training data is no longer available.

ViT-Up achieves the best $\mathcal{J}\&\mathcal{F}$ for both backbones and protocols. On DINOv2-S, it exceeds the strongest baselines by $2.2$ points in probing and $2.1$ in k-NN propagation. Notably, LILA trains on YouTube-VOS~\cite{18_youtube_vos} videos with optical-flow and depth supervision to obtain temporally consistent features, and flow cues add $1.4$ k-NN points in its ablation. ViT-Up learns only from still ImageNet-1K images, yet outperforms LILA under both protocols, and its k-NN accuracy degrades over time at a similar rate ($\mathcal{J}$ decay $12.4$ versus $11.5$). Faithful upsampling of backbone features thus appears to provide much of the consistency that LILA learns from motion.

On DINOv3-S+, the margins widen to $3.5$ and $5.7$ points, while NAF falls below bilinear interpolation in k-NN. The supplement reports other grids and all DAVIS metrics.

\subsection{Ablation Studies}

\begin{table}[t]
\centering
\footnotesize
\setlength{\cw}{0.78cm}
\setlength{\aboverulesep}{0pt}
\setlength{\belowrulesep}{0pt}
\renewcommand{\arraystretch}{1.3}
\setlength{\tabcolsep}{2pt}

\caption{DINOv3-S+ ablations: segmentation mIoU and SPair-71k PCK (\%, higher is better). Base is six-layer Stage~1 with LoRA in the first three blocks and full Stage~2 in the last. Parameters count weights needed for final feature prediction, including LoRA before merging and the final output head, but excluding the pretrained backbone and auxiliary heads. Stage~2 adds none. Within each block, the top five scores per column are shaded darkest to lightest, with the best bold. Ties share a rank.}
\label{tab:ablation}

\begin{adjustbox}{max width=\columnwidth}
\begin{tabular}{l c *{5}{C}}
\toprule
& \textbf{Params}
& \multicolumn{2}{c}{\textbf{Seg.: mIoU$\uparrow$}}
& \multicolumn{3}{c}{\textbf{SPair-71k: PCK$\uparrow$}} \\
\cmidrule(lr){3-4}\cmidrule(lr){5-7}
\textbf{Variant} & \metric{(M)}
& \metric{VOC} & \metric{City.}
& \metric{0.10} & \metric{0.05} & \metric{0.01} \\
\midrule

\sect{Component ablations}
Base & 23.99
& \bs2{87.45} & \bs2{65.44} & \bs2{55.60} & \bs1{\textbf{39.43}} & \bs1{\textbf{7.75}} \\
Cross-Attention off & 20.44
& 86.53 & 62.57 & 52.13 & 35.34 & 6.48 \\
FeatX off & 14.96
& \bs3{87.44} & \bs5{65.22} & \bs5{55.40} & \bs4{38.89} & \bs3{7.42} \\
LoRA off & 23.38
& \bs5{86.97} & \bs1{\textbf{66.27}} & \bs3{55.46} & \bs3{39.28} & \bs4{7.38} \\
Decoder off & 23.84
& \bs1{\textbf{87.48}} & \bs4{65.28} & \bs1{\textbf{55.70}} & \bs2{39.33} & \bs2{7.46} \\
Transition off & 20.44
& \bs4{87.42} & \bs3{65.42} & \bs4{55.45} & \bs5{38.87} & \bs5{7.33} \\

\sect{Architectural variants}
Base & 23.99
& \bs1{\textbf{87.45}} & \bs2{65.44} & \bs2{55.60} & \bs1{\textbf{39.43}} & \bs2{7.75} \\
Learned Query Embd. & 23.99
& \bs3{87.32} & \bs1{\textbf{65.57}} & \bs1{\textbf{55.62}} & \bs1{\textbf{39.43}} & \bs1{\textbf{7.87}} \\
\hspace{0.8em}+ L0 Conditioning & 26.68
& \bs2{87.33} & \bs3{65.36} & \bs3{55.47} & \bs4{39.19} & \bs3{7.74} \\
Parallel Blocks & 25.02
& \bs5{87.05} & \bs4{64.22} & 54.37 & \bs5{37.77} & \bs5{6.74} \\
Last State Repeated & 23.99
& \bs4{87.13} & 63.88 & \bs5{54.68} & 37.42 & 6.04 \\
\hspace{0.8em}+ Interm. Sup. & 23.99
& 85.77 & 62.90 & \bs4{55.17} & \bs3{39.30} & \bs4{7.36} \\
Shared Block Weights & 4.63
& 86.84 & \bs5{64.05} & 54.13 & 37.28 & 6.33 \\

\sect{Refinement layers}
1 layer & 4.63
& \bs5{86.64} & \bs5{63.18} & \bs5{53.35} & \bs5{35.35} & \bs5{4.93} \\
2 layers & 8.50
& \bs3{87.34} & \bs4{64.80} & \bs4{54.34} & \bs4{37.26} & \bs4{6.62} \\
3 layers & 12.37
& \bs4{87.27} & \bs2{65.45} & \bs3{54.76} & \bs3{38.26} & \bs3{7.43} \\
6 layers (base) & 23.99
& \bs1{\textbf{87.45}} & \bs3{65.44} & \bs2{55.60} & \bs2{39.43} & \bs2{7.75} \\
12 layers & 47.22
& \bs1{\textbf{87.45}} & \bs1{\textbf{65.65}} & \bs1{\textbf{55.92}} & \bs1{\textbf{39.71}} & \bs1{\textbf{7.86}} \\

\sect{Stage 2 ablations}
Base & 23.99
& \bs1{\textbf{87.59}} & \bs1{\textbf{66.42}} & \bs2{56.39} & \bs1{\textbf{41.18}} & \bs2{8.98} \\
w/o anchor & 23.99
& \bs3{87.42} & \bs3{64.59} & \bs3{54.78} & \bs3{39.62} & \bs3{8.38} \\
w/o offset reg. & 23.99
& \bs2{87.57} & \bs2{66.37} & \bs1{\textbf{56.42}} & \bs2{41.10} & \bs1{\textbf{9.00}} \\
\bottomrule
\end{tabular}
\end{adjustbox}
\end{table}

All quantitative Stage~1 ablations use one ImageNet-1K epoch, and Stage~2 ablations use $20{,}000$ additional steps. Table~\ref{tab:ablation} evaluates components, architectural variants, refinement depth, and Stage~2 losses through segmentation probing and SPair-71k correspondence.

\paragraph{Individual Components}

\begin{figure*}[t]
    \centering
    \includegraphics[width=\textwidth]{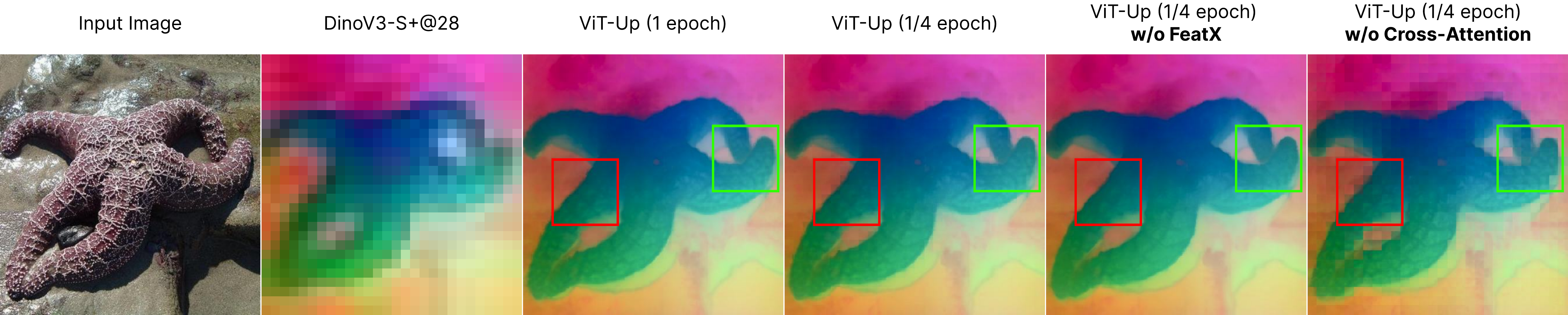}
    \caption{
        Qualitative ablation of cross-attention and FeatX in Stage~1 ViT-Up on DINOv3-S+.
        The input image has resolution $448{\times}448$, the native DINOv3-S+ feature grid has resolution $28{\times}28$, and all dense feature maps are shown at $448{\times}448$ resolution.
        From left to right, we show the input RGB image, DINOv3-S+@28, ViT-Up after one epoch of training, and the one-quarter-epoch ablated variants of ViT-Up in its base configuration, without FeatX, and without cross-attention.
        Without FeatX, the features remain spatially consistent but lose fine local texture, most visibly near the arm ends of the sea star highlighted by the green box.
        Without cross-attention, the features preserve more local detail but exhibit pronounced pixelation artifacts, especially in the region highlighted by the red box.
        Combining both components yields coherent features with visible local detail. The remaining pixelation artifacts after one-quarter epoch of training largely disappear after one full epoch. Best viewed digitally.
    }
    \label{fig:local_vs_global}
\end{figure*}

Cross-attention has the largest effect: removing it lowers every metric, most strongly Cityscapes and correspondence. FeatX contributes less numerically, but its removal also lowers all metrics, with the clearest losses at strict PCK thresholds. Fig.~\ref{fig:local_vs_global} explains their complementary roles: without FeatX, sea-star features remain coherent but lose fine texture. Without attention, local detail survives but pixelation appears. Combining both yields coherent detail, with residual quarter-epoch pixelation largely disappearing after a full epoch.

As discussed in Sec.~\ref{sec:dense_linear_probing}, LoRA improves VOC and correspondence but lowers Cityscapes mIoU. Removing the output decoders slightly improves VOC and PCK@0.10, but lowers Cityscapes and PCK@0.05/0.01. Direct intermediate-state supervision may constrain query representations toward the backbone's semantics while limiting their capacity to retain high-frequency detail. Removing the transition MLP leaves VOC and Cityscapes nearly unchanged but lowers PCK at every threshold. Without this mapping across skipped backbone layers, semantic reconstruction errors may accumulate through refinement. Together, these results suggest that output decoders preserve flexibility for spatial detail, while transitions help maintain semantic alignment across layers.

\paragraph{Architectural Variants}

The second block of Table~\ref{tab:ablation} separates query initialization, intermediate-state access, progressive processing, supervision, and parameter sharing.

Replacing the image-derived initializer with a learned vector leaves every metric within $0.13$ points of the base model, with no consistent advantage. Position remains available through FeatX and continuous RoPE. Adding native-grid embedding-layer features through FeatX and a residual fusion MLP (L0 Conditioning) adds $2.69$M parameters without a consistent benefit. Query initialization is therefore a design choice. We retain the shared patch embedding because it starts queries in the backbone's embedding space without adding parameters. At $448^2$ output resolution, its overhead over learned queries is only $3.14$~ms, or $6.8\%$ of total runtime.

Intermediate hidden states matter independently of parameter count. Feeding the final backbone state to every block while retaining layer-wise supervision (Last State Repeated + Interm.\ Sup.) lowers all metrics at identical capacity. Supervising only the final output (Last State Repeated) also remains below the base model. Access to intermediate backbone states therefore improves both segmentation and correspondence.

Progressive processing also matters beyond multi-layer access. In Parallel Blocks, each block receives the same initial query and its assigned backbone state. Their outputs are summed and passed through a residual output MLP. Only the final output is supervised, since there is no intermediate refinement sequence. Despite access to the same backbone states and nearly matched capacity ($25.02$M versus $23.99$M), this variant performs worse on every metric, supporting progressive refinement of the query representation.

Sharing one block across all six refinement steps while retaining the output decoders reduces parameters from $23.99$M to $4.63$M. This compact model still outperforms the baselines on VOC, Cityscapes, and PCK@0.10/0.05, and all except DPT at PCK@0.01 (Tables~\ref{tab:probing} and~\ref{tab:correspondence}). It trails the base model on every metric, supporting layer-specific block parameters. Weight sharing reduces storage, but retains all six block applications and therefore does not itself reduce refinement computation.

\paragraph{Number of Refinement Layers}
Increasing refinement depth consistently improves correspondence, particularly at stricter PCK thresholds. Three layers already outperform DPT on all five metrics with fewer parameters ($12.37$M versus $18.70$M). Most of the benefit is reached at six layers: relative to twelve, VOC is tied, Cityscapes is within $0.21$ mIoU, and every PCK score is within $0.32$ points, with roughly half the parameters. We therefore use six layers to balance accuracy and cost.

\paragraph{Stage 2 Optimization}
The weak anchor loss is essential to refinement: removing it lowers Cityscapes and every PCK score, with Cityscapes and PCK@0.10 falling below Stage~1. This supports retaining unshifted targets to preserve spatial structure while sharpening. Offset regularization has a much smaller effect, slightly benefiting segmentation and giving mixed correspondence changes. We retain it to discourage large displacements at negligible computational cost.

\subsection{Runtime and Memory}
\label{sec:runtime_memory}

\begin{table}[t]
\centering
\scriptsize
\setlength{\cw}{0.8cm}
\setlength{\aboverulesep}{0pt}
\setlength{\belowrulesep}{0pt}
\renewcommand{\arraystretch}{1.3}
\setlength{\tabcolsep}{2pt}

\caption{Runtime and peak allocated memory on an H100 SXM GPU (700~W), including DINOv3-S+. Parameter counts follow Table~\ref{tab:ablation}. Chunk is the number of queries processed together. Lower is better. The top five scores per metric are shaded darkest to lightest, with the best bold. Dashes denote omitted or inapplicable measurements.}
\label{tab:runtime_memory}

\begin{adjustbox}{max width=\columnwidth}
\begin{tabular}{l c c *{6}{C}}
\toprule
& & & \multicolumn{3}{c}{\textbf{Time [ms]$\downarrow$}}
& \multicolumn{3}{c}{\textbf{VRAM [MiB]$\downarrow$}} \\
\cmidrule(lr){4-6}\cmidrule(lr){7-9}
\textbf{Method} & \textbf{Params} & \textbf{Chunk}
& \metric{$112^2$} & \metric{$224^2$} & \metric{$448^2$}
& \metric{$112^2$} & \metric{$224^2$} & \metric{$448^2$} \\
\midrule
JAFAR & 0.6M & --
& 42.1 & 44.3 & 54.2
& \bs5{600.2} & 1781.7 & 6508.1 \\
AnyUp & 0.8M & --
& \bs3{7.8} & \bs4{16.7} & 63.3
& \bs4{545.9} & \bs5{833.8} & 2821.6 \\
UpLiFT & 0.8M & --
& \bs2{7.1} & \bs1{\textbf{7.8}} & \bs2{10.7}
& \bs1{\textbf{261.5}} & \bs1{\textbf{420.1}} & \bs3{1045.9} \\
NAF & 0.7M & --
& \bs5{30.2} & 31.9 & \bs4{38.7}
& 662.9 & \bs4{663.2} & 1840.2 \\
LoftUp & 4.3M & --
& \bs1{\textbf{5.0}} & \bs2{9.1} & \bs3{25.4}
& \bs2{268.2} & \bs2{520.4} & \bs5{1523.3} \\
DPT & 18.7M & --
& -- & -- & \bs1{\textbf{9.4}}
& -- & -- & \bs1{\textbf{849.0}} \\

\arrayrulecolor{rulec}\midrule[0.4pt]\arrayrulecolor{black}

\multirow{3}{*}{\textbf{ViT-Up}} & \multirow{3}{*}{24.0M} & $112^2$
& \bs4{8.8} & \bs5{19.9} & 63.2
& \bs3{545.4} & \bs3{656.8} & \bs2{851.1} \\
& & $224^2$
& -- & \bs3{16.4} & 49.8
& -- & 1189.7 & \bs4{1300.0} \\
& & $448^2$
& -- & -- & \bs5{46.1}
& -- & -- & 3316.8 \\
\bottomrule
\end{tabular}
\end{adjustbox}
\end{table}

Table~\ref{tab:runtime_memory} measures full forward passes, including backbone extraction and dense output, on an H100 SXM GPU (700~W), with batch size $1$ and $448^2$ inputs. We use BF16 autocast with FP32 weights, inputs, coordinates, and outputs, and merge LoRA weights. Runtime is the median of five repeat medians. Peak allocated memory includes the model, inputs, and outputs. Parameter counts follow Table~\ref{tab:ablation}. Timings and memory include retained auxiliary heads. The supplement specifies warmup and measurement settings.

LoftUp uses its released weights with DINOv3-S+ only to measure computational cost, not task performance. DPT is reported only at its fixed $448^2$ output. UpLiFT uses the optimized compiled implementation at recursion-aligned resolutions. Compilation takes approximately $30$--$60$ seconds before inference.

ViT-Up's runtime depends strongly on the requested output resolution. At $448^2$, it is faster than JAFAR and AnyUp, but slower than NAF, LoftUp, UpLiFT, and DPT. At $112^2$, it takes $8.8$~ms, only $1.7$~ms more than UpLiFT, while outperforming prior upsamplers evaluated at full resolution in trained probing (Table~\ref{tab:vitup_resolution_probing}). NAF and JAFAR retain the cost of encoding the full guidance image before pooling, limiting their benefit from a smaller output grid.

Conditionally independent queries permit an explicit memory--latency trade-off through chunking without changing the output. At $448^2$ output, $112^2$ query chunks reduce peak memory from $3316.8$ to $851.1$~MiB, close to DPT's $849.0$~MiB, while increasing runtime from $46.1$ to $63.2$~ms. Query independence also allows sparse evaluation or dense sampling within a region of interest, potentially providing substantial speedups. We leave exploration of these inference modes to future work.

\section{Limitations and Outlook}
\label{sec:limitations}

\paragraph{Information Bottleneck}
ViT-Up's reconstruction of backbone-aligned semantics depends on the information retained across the backbone hierarchy. Intermediate states provide spatial and semantic evidence beyond the final token grid, but fine-grained distinctions that are weakly represented throughout the hierarchy can remain difficult to recover.

One practical solution is to increase the backbone grid from $28\times28$ to $56\times56$, providing ViT-Up with richer spatial information. Feature quality is preserved while quadratic attention remains tractable. The improvements on the $56\times56$ grid (Sec.~IV.H) show that higher backbone resolution and learned upsampling are complementary.

\paragraph{Backbone Coupling}
ViT-Up requires training for each target backbone checkpoint. Transition MLPs align skipped hidden layers, and FeatX extracts sub-token information from their representations. Both depend on how that backbone organizes spatial and semantic content. Once trained, the decoder is reusable across downstream tasks, datasets, and output resolutions for the same backbone.

A longer-term direction is to train the backbone and decoder jointly from scratch, making continuous high-resolution feature querying a native capability. High-resolution reconstruction losses could then shape the hidden states directly, encouraging them to retain local detail that current backbone objectives may discard. Our LoRA adaptation offers a limited step toward this joint optimization.

\section{Conclusion}
\label{sec:conclusion}

We introduced ViT-Up, an implicit feature upsampling framework that replaces external image guidance with progressive query construction from intermediate ViT representations. By following the backbone hierarchy, ViT-Up recovers spatial detail and semantic structure beyond the native token grid. Stage~1 learns dense feature reconstruction from multi-scale backbone targets, while Stage~2 reduces inherited blur and leakage through offset-guided self-distillation.

ViT-Up outperforms existing upsamplers in dense linear probing and semantic correspondence, and scales favorably with backbone capacity. Its coordinate-conditioned formulation allows output density to be chosen independently of the backbone grid, with useful additional detail even at high output resolutions.

These results demonstrate the value of the ViT hierarchy for faithful feature upsampling and motivate future backbones that support high-resolution feature querying directly.

\section*{Acknowledgments}
We thank Nils Wandel for proofreading the manuscript. This work was supported in part by the National Natural Science Foundation of China under Grants 62225309, U24A20278, and 62361166632. The authors used GitHub Copilot, OpenAI Codex, and ChatGPT for implementation support and language editing. All scientific content was developed and verified by the authors.

\bibliographystyle{IEEEtran}
\bibliography{main}

\clearpage
\twocolumn[{\centering{\LARGE Supplementary Material}\par\vspace{1.5em}}]
\appendices

\section{Extended Qualitative Analysis}
\label{app:extended_qualitative}

\subsection{DINOv2 Artifacts}
\label{app:dinov2_artifacts}

\begin{figure*}[t]
    \centering
    \includegraphics[width=\textwidth]{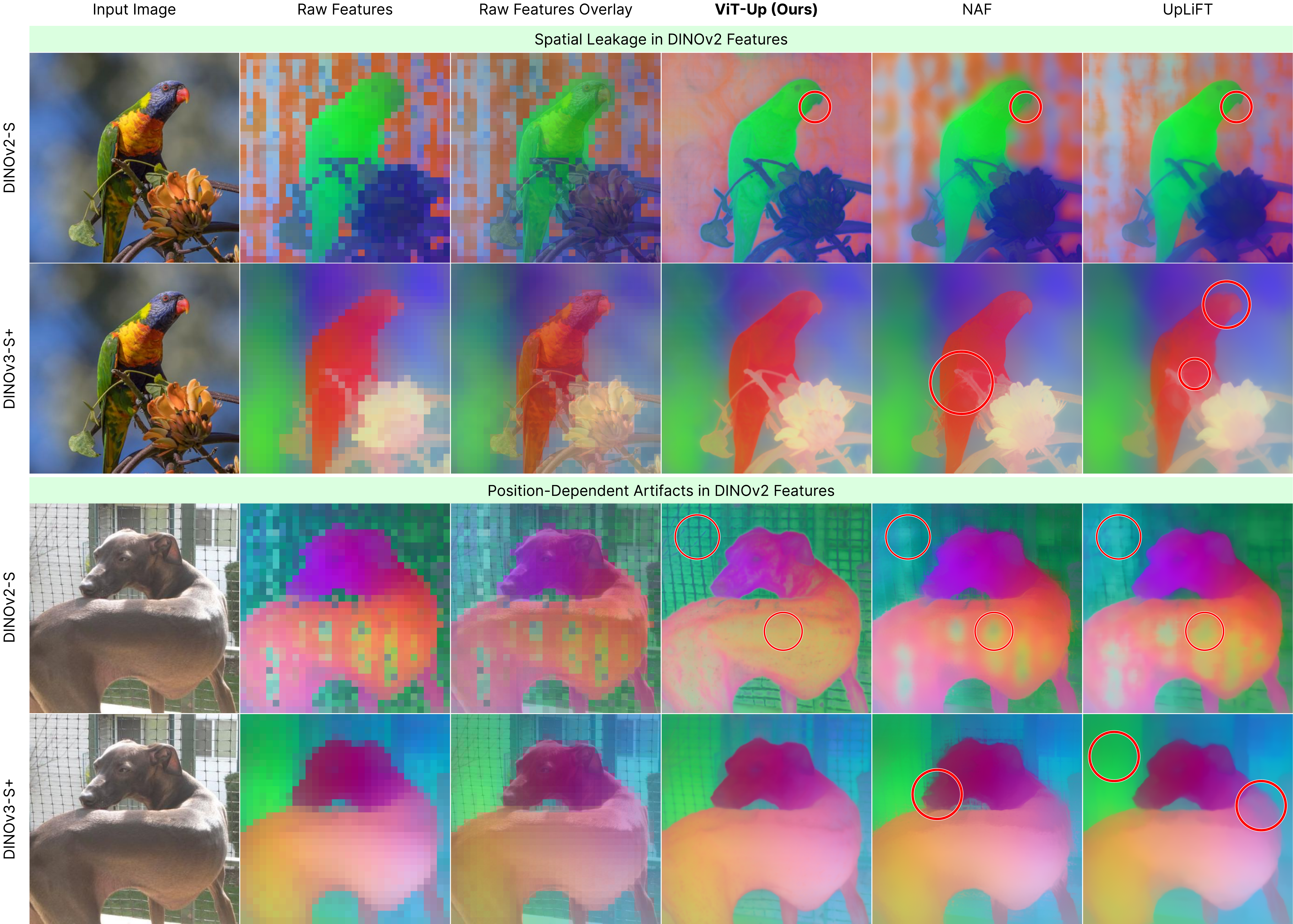}
    \caption{
    DINOv2 feature artifacts.
    For each example, the top row shows DINOv2-based features and the bottom row shows DINOv3-based features under the same visualization protocol.
    DINOv2 exhibits stronger spatial leakage and position-dependent artifacts, while DINOv3 provides a cleaner dense feature field.
    ViT-Up uses Stage~2 on both backbones, with crop supervision in Stage~1 for DINOv2-S.
    PCA is fitted to multiscale raw backbone features and shared across methods within each image and backbone.
    DINOv2 circles mark leakage near the bird's head, fence detail in the dog's background, and positional artifacts within its body, at identical coordinates across methods.
    }
    \label{fig:artifacts}
\end{figure*}

DINOv2 features exhibit both position-dependent patterns and spatial leakage across object boundaries. These artifacts complicate feature upsampling because a reconstruction objective can reproduce undesired structure already present in the supervision signal. We examine these effects qualitatively, then separate the contributions of crop supervision and Stage~2 refinement. All examples use the same square image crops and $448\times448$ output grids.

\paragraph{Visualization protocol}
For each image and backbone, we extract the final-layer features of the frozen, pretrained backbone at token-grid sizes $32\times32$, $64\times64$, and $112\times112$. We L2-normalize the feature vectors, concatenate the tokens across scales, and fit a three-component PCA after centering, without channel-wise standardization. The resulting projection and color limits are fixed for all upsampled feature maps of that image and backbone. Thus, neither the upsampling method nor the training stage influences the PCA fit. Colors are comparable within each backbone comparison, but not across backbones. The raw-feature panels show the $32\times32$ reference grid with nearest-neighbor enlargement.

\paragraph{Position artifacts and feature leakage}
Fig.~\ref{fig:artifacts} illustrates two distinct deficiencies. In the bird example, the DINOv2 feature field extends beyond the object boundary near the head, mixing foreground and background features. In the dog example, repeated spatial patterns appear within the body even though they do not follow its semantic structure. Such position-dependent patterns have also been studied in vision-transformer denoising~\cite{24_denoising_vit}. DINOv3 provides a cleaner reference in both examples.

The DINOv2 comparison uses ViT-Up with crop supervision and Stage~2 refinement. NAF and UpLiFT produce smoother maps, but repeated patterns remain within the dog's body and in the bird's background. ViT-Up suppresses much of this structure while retaining the object contours and the branches around the bird. The following comparisons distinguish the effect of changing the Stage~1 supervision from the subsequent refinement.

\subsection{Stage 2 Refinement on DINOv2 and DINOv3}
\label{app:stage2_qualitative}
\begin{figure*}[t]
    \centering
    \includegraphics[width=\textwidth]{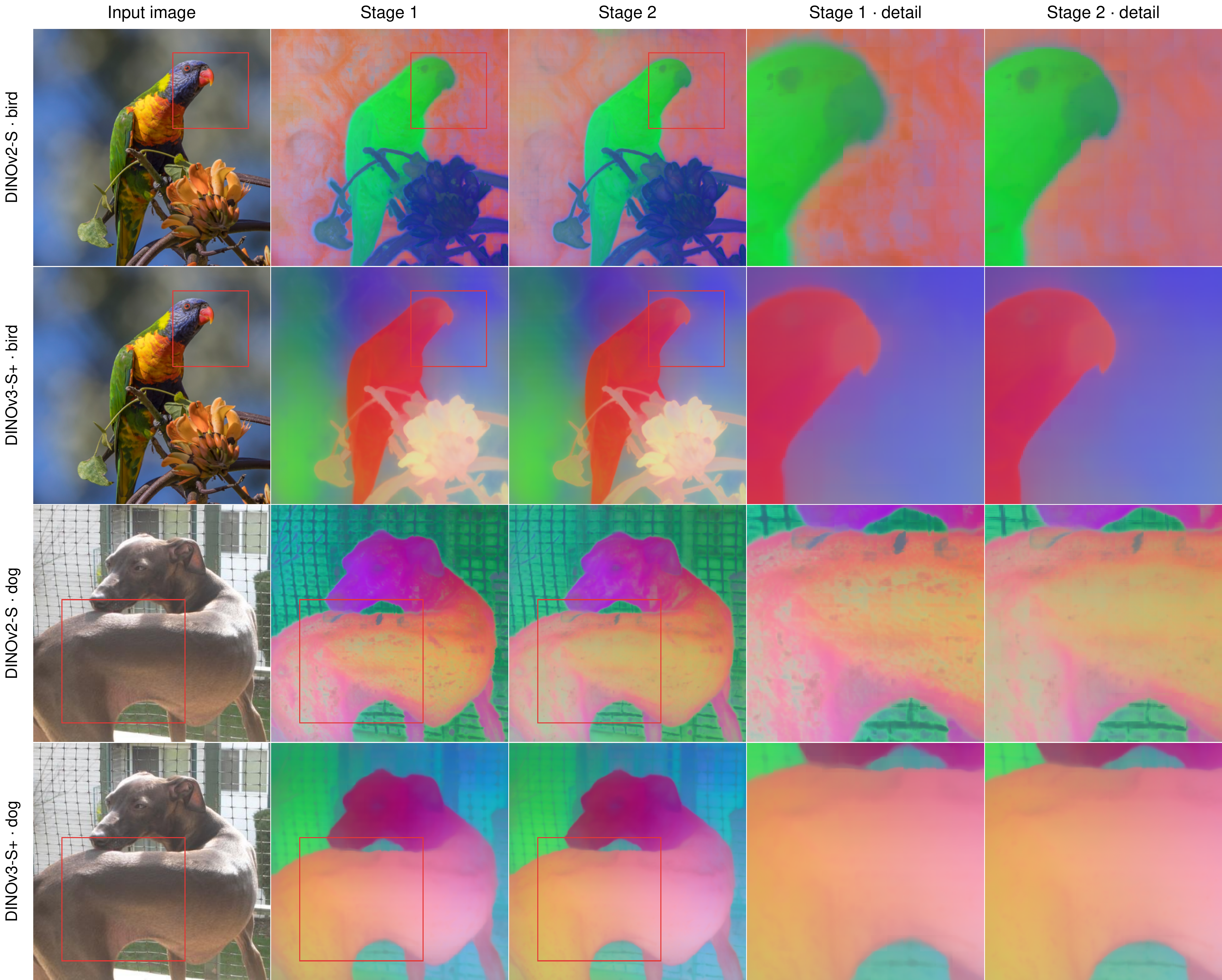}
    \caption{Stage~2 refinement on DINOv2-S and DINOv3-S+. Each row compares a Stage~1 model with its Stage~2 refinement using the same input and a shared PCA projection fitted only to multiscale raw backbone features. The rightmost columns enlarge the regions marked in red. These regions expose feature leakage and blur around the bird's head and residual spatial variation within the dog's body.}
    \label{fig:stage2_refinement}
\end{figure*}

Fig.~\ref{fig:stage2_refinement} compares each Stage~1 model with its Stage~2 refinement on DINOv2-S and DINOv3-S+. Each pair uses identical inputs, output coordinates, and the same raw-backbone PCA projection. The DINOv2 pair uses crop supervision in Stage~1; the DINOv3 pair includes the Stage~1 LoRA adaptation. The backbone is fixed during Stage~2.

Stage~2 reduces feature leakage and blur that remain after reconstruction training. On DINOv2, the transition around the bird's beak becomes more localized, and the fine mottled variation within the dog's body is attenuated while its contour is retained. The stronger leakage in the initial DINOv2 representation gives refinement more scope to improve spatial localization. DINOv3 starts from cleaner features, and its boundary and intra-object changes are subtler in these examples. All comparisons retain the same color scale so that differences in contrast do not arise from separately normalizing the two stages.

\begin{figure*}[t]
    \centering
    \includegraphics[width=\textwidth]{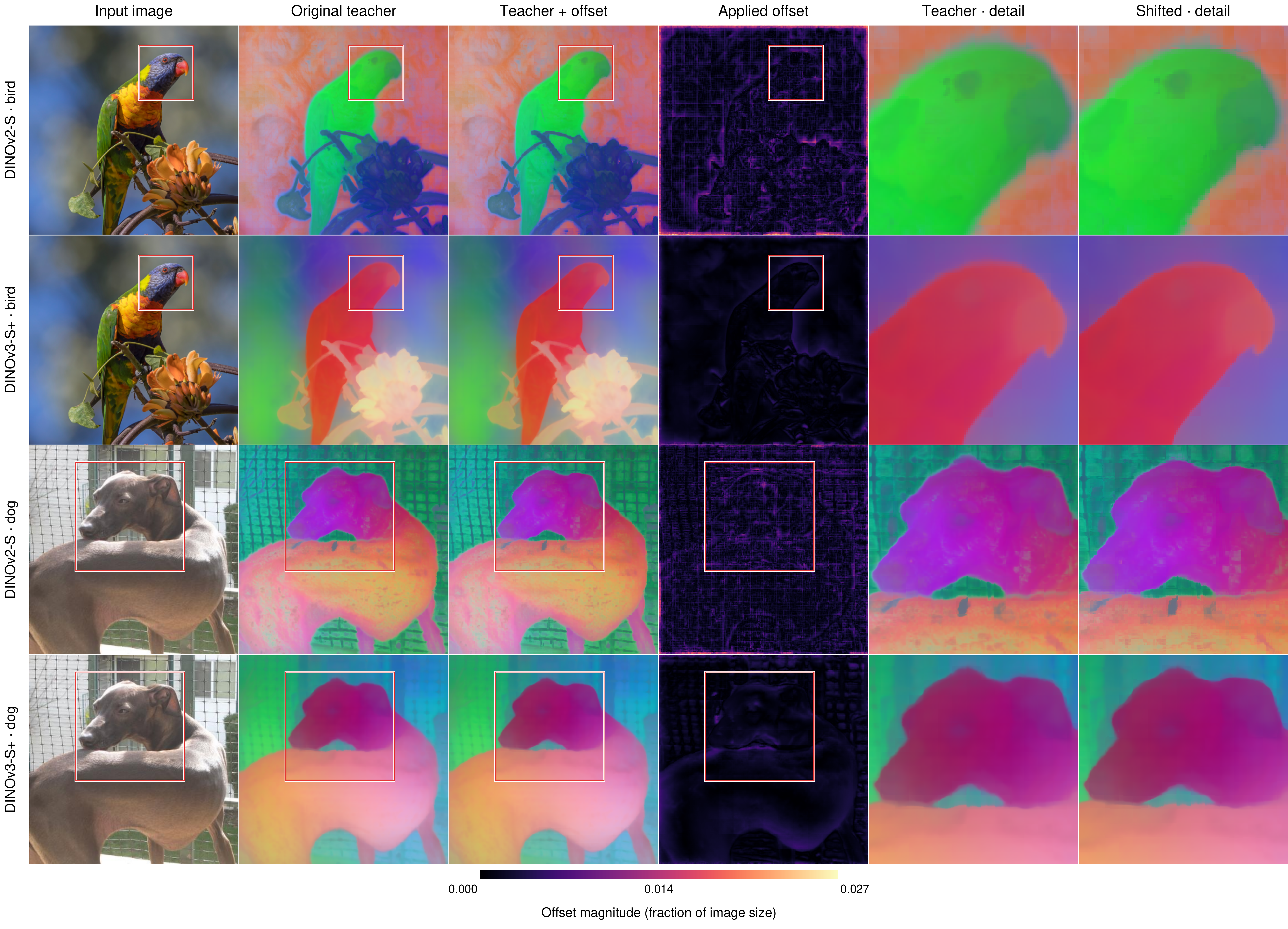}
    \caption{Learned Stage~2 offsets on DINOv2-S and DINOv3-S+. The original frozen Stage~1 teacher is queried at $\mathbf q$ and at $\operatorname{clip}(\mathbf q+\boldsymbol\Delta(\mathbf q),0,1)$; both feature maps are displayed at the original locations $\mathbf q$ using the same raw-backbone PCA projection and color limits. The student predicts offsets from a centered scale-$0.5$ view, while the teacher receives the full image. Heatmaps show the applied displacement magnitude after clipping to the image domain, measured in normalized full-image coordinates, with one scale shared across all rows. The rightmost columns enlarge the regions marked in red, using identical crops and color scales before and after applying the offsets.}
    \label{fig:stage2_offsets}
\end{figure*}

\paragraph{Learned coordinate offsets}
Fig.~\ref{fig:stage2_offsets} isolates the effect of applying the learned offsets to the original frozen Stage~1 teacher. For each backbone, the Stage~2 student receives the image downscaled to $224\times224$ and centered in a $448\times448$ canvas with the training noise background; the teacher receives the full $448\times448$ image. Student queries are mapped into the padded content region, while offsets are expressed in the full-image coordinate system. We evaluate the teacher directly at both the original and shifted coordinates on a $448\times448$ output grid. The shifted feature maps therefore show the adjusted training targets, rather than the final student's predictions or a warp of a precomputed feature map. The heatmaps show the applied displacement after clipping shifted queries to the image domain, with one color scale shared across both images and backbones. Matched zooms expose the changes within the marked regions. Each teacher checkpoint matches the initialization recorded by its Stage~2 training run.

\paragraph{Coordinate gradients}
Stage~2 learns the two offset components by differentiating the feature-matching loss through the frozen teacher's query coordinates. Freezing the teacher's parameters preserves this coordinate-gradient path. A piecewise-constant feature field would yield zero coordinate gradients almost everywhere and could not guide offset learning through that loss. Successful offset optimization thus provides evidence of useful local coordinate dependence, complementing the output-resolution experiment. Discrete image pixels and backbone tokens do not preclude learning a continuous feature field.

\subsection{Crop versus Uniform Supervision on DINOv2}
\label{app:crop_supervision}
\begin{figure*}[t]
    \centering
    \includegraphics[width=\textwidth]{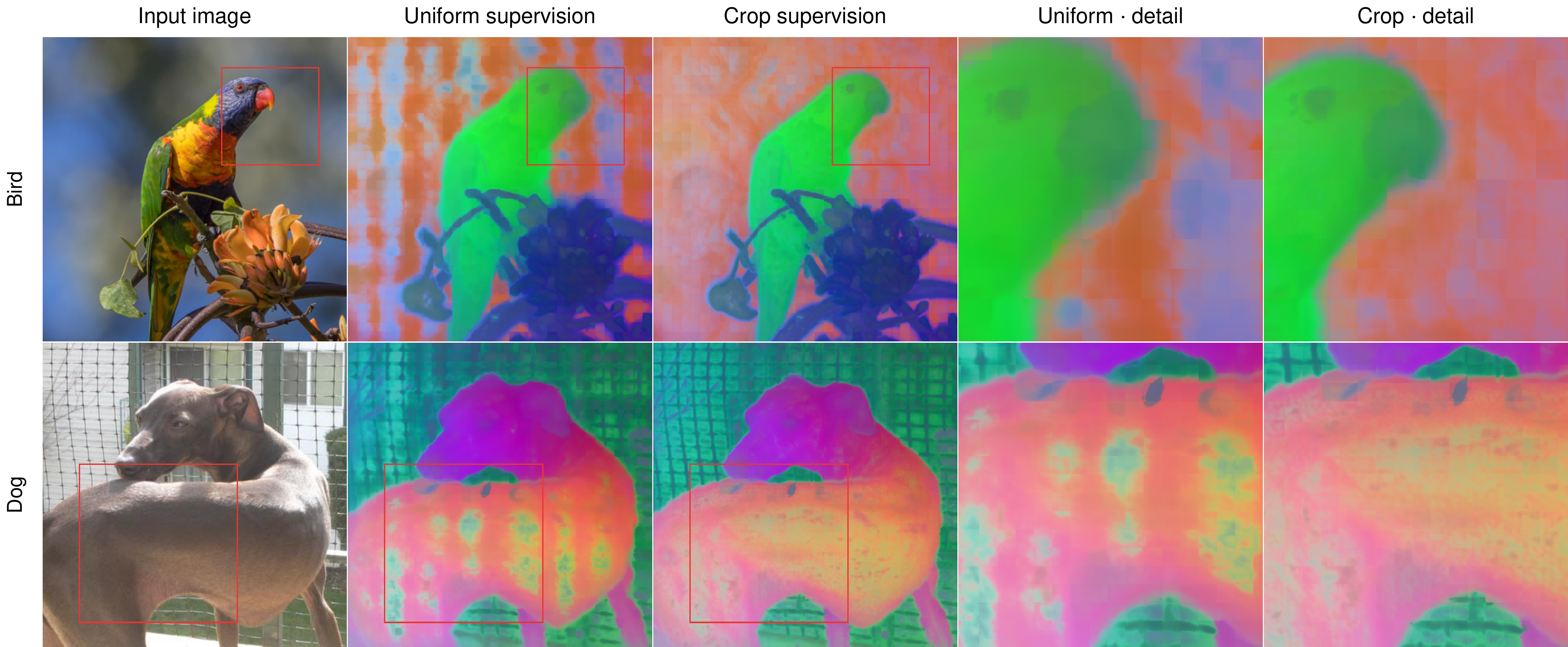}
    \caption{Crop versus uniform supervision for DINOv2-S Stage~1. Both models use the same architecture and training budget, with checkpoints selected by validation feature loss. The PCA basis and color limits are fitted to multiscale raw DINOv2 features and applied unchanged to both outputs. The rightmost columns show the same enlarged regions. This comparison precedes Stage~2 and isolates the change in Stage~1 supervision.}
    \label{fig:crop_supervision}
\end{figure*}

Fig.~\ref{fig:crop_supervision} compares two DINOv2-S Stage~1 models trained with the same architecture and a budget of $20{,}000$ optimizer updates, selecting each checkpoint by validation feature loss. Uniform supervision uses the full-image teacher while varying the student's content scale and position. Crop supervision combines full-image targets with teacher features extracted from half-size crops, with the student and teacher coordinates aligned over the corresponding image region. Neither model in this comparison uses Stage~2.

The uniform-supervision model reproduces pronounced vertical bands in the bird's background and repeated spots across the dog's body. Crop supervision substantially weakens both patterns, making features within object regions more coherent. It does not by itself eliminate leakage or blur at object boundaries; these are the focus of the subsequent Stage~2 refinement. The two comparisons therefore examine complementary changes: the supervision used to learn the initial feature field and the refinement of its spatial structure.

\section{Experimental Details}
\label{app:experimental_details}

This section specifies the DINOv3 training and evaluation settings used in our experiments. We distinguish model training, task-specific probing, frozen-head evaluation, and inference benchmarks.

\subsection{Model Architecture}

\begin{table*}[t]
\centering
\small
\caption{DINOv3 model dimensions and architecture. Spatial sizes are side lengths of square images or feature grids.}
\label{tab:supp_model_details}
\begin{tabular}{p{0.40\textwidth} p{0.25\textwidth} p{0.25\textwidth}}
\toprule
Setting & DINOv3-S+ & DINOv3-B \\
\midrule
Backbone patch size & $16$ & $16$ \\
Feature and ViT-Up dimension $C$ & $384$ & $768$ \\
Backbone input / native grid & $448$ / $28$ & $448$ / $28$ \\
Refinement blocks / attention heads & $6$ / $6$ & $6$ / $12$ \\
Supervised backbone layers & $0,2,4,6,8,10,12$ & $0,2,4,6,8,10,12$ \\
Query-embedding input / patch grid & $3584$ / $224$ & $3584$ / $224$ \\
Transition MLP widths & $384,768,384$ & $768,1536,768$ \\
FeatX output and fusion MLP widths & $384,1536,384$ & $768,3072,768$ \\
FiLM MLP widths & $66,384,768$ & $66,768,1536$ \\
Output projection & LayerNorm + $384\rightarrow384$ & LayerNorm + $768\rightarrow768$ \\
LoRA rank / scale / dropout & $16$ / $32$ / $0.05$ & $16$ / $32$ / $0.05$ \\
\bottomrule
\end{tabular}
\end{table*}

We use the pretrained DINOv3-S+ and DINOv3-B backbones with patch size $16$, identified by \path{facebook/dinov3-vits16plus-pretrain-lvd1689m} and \path{facebook/dinov3-vitb16-pretrain-lvd1689m}, respectively. Table~\ref{tab:supp_model_details} gives the corresponding ViT-Up dimensions. Both models use six refinement blocks and supervise the embedding layer and backbone layers $2,4,6,8,10,12$. Block parameters and output projections are independent across layers unless weight sharing is explicitly ablated.

The transition MLP has widths $C\rightarrow2C\rightarrow C$, and the FeatX output MLP and fusion MLP have widths $C\rightarrow4C\rightarrow C$, where $C$ is the backbone feature dimension. Hidden layers use GELU. The FeatX positional input concatenates the two raw relative coordinates with sine and cosine features from $16$ frequency bands per coordinate, giving $66$ channels in total. Frequency band $j$ is $10^{j/15}$ for $j=0,\ldots,15$, with a factor of $\pi$ in the sine/cosine arguments. Its FiLM MLP has widths $66\rightarrow C\rightarrow2C$ and a zero-initialized final projection. Each output decoder consists of LayerNorm followed by a $C\rightarrow C$ linear projection. Cross-attention uses a configured window side of $32$ tokens, biased projections, and zero attention and projection dropout. This covers the full $28^2$ backbone grid. The implementation also falls back to full attention when query and backbone grids differ. Continuous RoPE uses $64$ channels per head, base $100$, and coordinate scale $2\pi$.

Stage~1 adapts the backbone patch embedding and attention query, key, value, and output projections using LoRA with rank $16$, scale $32$, and dropout $0.05$; backbone biases and the original backbone weights remain frozen. The no-LoRA control removes these adapters. Stage~2 freezes the inherited Stage~1 backbone, including any LoRA adapters.

\subsection{Training Protocol}

\begin{table*}[t]
\centering
\small
\caption{Final ViT-Up training settings. Batch sizes are per GPU; accumulation gives effective batches of $24$ and $16$. Both stages use one GPU and BF16 mixed precision.}
\label{tab:supp_training_details}
\begin{tabular}{p{0.38\textwidth} p{0.27\textwidth} p{0.27\textwidth}}
\toprule
Setting & Stage 1 & Stage 2 \\
\midrule
Optimizer / weight decay & AdamW / $0$ & AdamW / $0$ \\
AdamW $(\beta_1,\beta_2)$ / $\epsilon$ & $(0.9,0.999)$ / $10^{-8}$ & $(0.9,0.999)$ / $10^{-8}$ \\
Gradient norm clipping & $1.0$ & $1.0$ \\
Peak / minimum learning rate & $2\times10^{-4}$ / $10^{-6}$ & $2\times10^{-4}$ / $10^{-6}$ \\
Optimizer updates & $53,381$ & $20,000$ \\
Linear warmup updates & $2,669$ & $1,000$ \\
Cosine decay updates & $50,712$ & $19,000$ \\
Initial learning-rate factor & $0.01$ & $0.01$ \\
DINOv3-S+ batch $\times$ accumulation & $24\times1$ & $16\times1$ \\
DINOv3-B batch $\times$ accumulation & $12\times2$ & $8\times2$ \\
Student content scale & Uniform $[0.1,1.0]$ & Fixed $0.5$ \\
Teacher & Frozen pretrained backbone & Frozen Stage 1 ViT-Up \\
Student backbone adaptation & Trainable LoRA & Frozen, including inherited LoRA \\
Queries per image & Regular $56\times56$ grid & $3,136$ random coordinates \\
Queries per training group & $3,136$ ($196$ blocks of $4^2$) & $3,136$ \\
Query activation checkpointing & Disabled & Disabled \\
Supervised layers / scales & Seven layers / $14,28,56$ & Final layer / one scale \\
L2 / cosine loss weight & $1$ / $1$ & $1$ / $1$ \\
Shifted / anchor / offset weights & Not applicable & $1$ / $0.1$ / $10$ \\
Training seed & $0$ & $0$ \\
\bottomrule
\end{tabular}
\end{table*}

\begin{table}[t]
\centering
\footnotesize
\caption{Training augmentations at strength $1.0$. Application probabilities account for nested random transforms. Color jitter and photometric distortion share an outer probability of $0.5$; distortion additionally uses its internal $p=0.5$. JPEG qualities have probabilities $(0.25,0.35,0.25,0.15)$.}
\label{tab:supp_augmentation_details}
\begin{tabular}{p{0.55\columnwidth} p{0.32\columnwidth}}
\toprule
Operation & Setting \\
\midrule
Color jitter: brightness, contrast, saturation, hue & $0.2,0.2,0.3,0.05$ \\
Photometric distortion & Internal $p=0.5$ \\
Grayscale & $p=0.1$ \\
Gaussian blur & Kernel $5$; $\sigma\in[0.3,1.5]$; $p=0.35$ \\
JPEG compression & Quality $95,80,60,40$ \\
Autocontrast / equalization & $p=0.1$ / $p=0.075$ \\
Sharpness adjustment & Factor $1.5$; $p=0.1$ \\
Posterization & $5$ bits; $p=0.05$ \\
Solarization & Threshold $0.9\times255$; $p=0.025$ \\
Inversion & $p=0.01$ \\
Gaussian noise & $\sigma=0.03$; $p=0.6$; clipped to $[0,1]$ \\
\bottomrule
\end{tabular}
\end{table}

\paragraph{Data and preprocessing}
We train on the ImageNet-1K training split and use its validation split for the feature-reconstruction validation loss. The source image is converted to RGB, resized with bilinear interpolation so that its shorter side is $896$, and center-cropped to $896\times896$. We do not apply random source cropping. Photometric augmentation is applied to the shared source before constructing the teacher and student views; its settings are listed in Table~\ref{tab:supp_augmentation_details}. Validation disables these photometric augmentations. Images use channel means $(0.485,0.456,0.406)$ and standard deviations $(0.229,0.224,0.225)$. As background augmentation, we add independent uniform RGB noise in $[0,1]$ to the unused pixels of the black canvas before normalization. The background and image content use the same normalization; background noise is also retained for feature-reconstruction validation.

\paragraph{Coordinate convention and view mapping}
Coordinates are ordered as $(x,y)$ and normalized to image boundaries $[0,1]^2$. Pixel $(i,j)$ in an image of height $H$ and width $W$ has center $((j+\tfrac12)/W,(i+\tfrac12)/H)$. Conversely, a normalized query $\mathbf q$ corresponds to pixel-index coordinates $(Wq_x-\tfrac12,Hq_y-\tfrac12)$. Bilinear feature sampling uses $2\mathbf q-1$ with \texttt{align\_corners=False} and border padding. Resizing a view therefore preserves its normalized coordinates.

The source below is the shared image after preprocessing. Let $S$ be its side length and the student canvas size ($S=896$ for DINOv3). A sampled student scale $s$ gives an integer content side $m=\max(1,\lfloor sS\rfloor)$. The resized source is pasted at an integer offset $\mathbf p=(p_x,p_y)$, with each component sampled uniformly from $\{0,\ldots,S-m\}$. Its normalized top-left and bottom-right bounds in the canvas are $\mathbf a=\mathbf p/S$ and $\mathbf b=(\mathbf p+(m,m))/S$.

For crop-supervised variants, let $\mathbf c,\mathbf d$ be the teacher crop's normalized bounds in the shared source, computed from its actual integer pixel edges. A crop-local teacher query $\mathbf q$ maps to the source and student as
\begin{align}
\mathbf u &= \mathbf c+(\mathbf d-\mathbf c)\odot\mathbf q, \\
\mathbf q_s &= \mathbf a+(\mathbf b-\mathbf a)\odot\mathbf u.
\label{eq:supp_query_mapping}
\end{align}
Equivalently, the teacher region has student-canvas bounds $\mathbf A=\mathbf a+(\mathbf b-\mathbf a)\odot\mathbf c$ and $\mathbf B=\mathbf a+(\mathbf b-\mathbf a)\odot\mathbf d$, so $\mathbf q_s=\mathbf A+(\mathbf B-\mathbf A)\odot\mathbf q$. The inverse mapping is $\mathbf q=(\mathbf u-\mathbf c)\oslash(\mathbf d-\mathbf c)$, where $\oslash$ denotes componentwise division. Bounds use the realized pixel dimensions after rounding, and mapped queries are not snapped to a token grid. For the full-image teachers used in the main DINOv3 experiments, $\mathbf c=\mathbf0$ and $\mathbf d=\mathbf1$, reducing the mapping to $\mathbf q_s=\mathbf a+(\mathbf b-\mathbf a)\odot\mathbf q$. Resizing the student canvas to the $448^2$ backbone input leaves these normalized coordinates unchanged.

\paragraph{Stage 1}
The teacher is the frozen pretrained backbone. It processes the shared image at $224^2$, $448^2$, and $896^2$ pixels, producing token grids of $14^2$, $28^2$, and $56^2$. The student content scale is sampled uniformly from $[0.1,1.0]$, and its position within the canvas is sampled randomly. Teacher-grid centers $((j+\tfrac12)/56,(i+\tfrac12)/56)$ are mapped into the student using Eq.~\eqref{eq:supp_query_mapping}. ViT-Up predicts this $56\times56$ grid over the visible image region; non-overlapping $2\times2$ and $4\times4$ average pooling aligns predictions with the $28^2$ and $14^2$ teacher grids, respectively. The seven supervised layers and three scales all have unit weights. Target-normalized L2 and cosine losses each have weight $1$ and numerical stabilizer $\epsilon=10^{-5}$. We average each loss over its selected layers and scales before adding the two terms; L2 also averages over feature channels and query locations. No relational or feature-statistics regularizer is used during ViT-Up training.

\paragraph{Stage 2}
Both the student and the frozen ViT-Up teacher are initialized from the same Stage~1 checkpoint: the checkpoint with the lowest validation feature loss for DINOv3-S+, and the released pretrained checkpoint for DINOv3-B. The teacher sees the full image at $448\times448$, while the student content scale is fixed to $0.5$. We sample $3,136$ continuous query coordinates per image and supervise only the final-layer features. The shifted-target, unshifted-anchor, and offset-magnitude weights are $1$, $0.1$, and $10$, respectively. Teacher parameters remain fixed; no EMA update or teacher-target blending is used.

The offset head applies LayerNorm, a $C\rightarrow256$ linear layer, GELU, and a $256\rightarrow2$ linear layer. The final layer is initialized to zero. Each predicted coordinate offset is bounded by $0.15\tanh(\cdot)$ in normalized image coordinates. The student is evaluated at $\mathbf q_s$ from Eq.~\eqref{eq:supp_query_mapping}; the anchor teacher is queried at $\mathbf q$, and the shifted teacher at $\operatorname{clip}(\mathbf q+\Delta\mathbf q,0,1)$ in teacher-view coordinates. The squared-offset penalty is evaluated before clipping. Coordinate arithmetic uses FP32. The offset head is used only to construct training targets and does not modify the feature-inference path. Gradients with respect to shifted coordinates pass through bilinear query sampling, continuous RoPE, and FeatX modulation. The nearest-token lookup is recomputed at each coordinate, with no gradient through the discrete index.

\paragraph{Optimization and checkpoint selection}
Table~\ref{tab:supp_training_details} gives the full schedules, measured in optimizer updates. Both stages use AdamW with learning rate $2\times10^{-4}$, $(\beta_1,\beta_2)=(0.9,0.999)$, optimizer $\epsilon=10^{-8}$, and zero weight decay. Linear warmup starts at $0.01$ times the peak learning rate and is followed by cosine decay to $10^{-6}$. Training uses one GPU, BF16 mixed precision, and seed $0$. The default validation interval is $500$ training batches with $10$ validation batches per check; the best checkpoint is selected by minimum validation feature loss. Checkpoint selection for the upsampler is separate from the final-epoch selection used for task-specific probing heads.

\paragraph{Query processing and gradient accumulation}
In the final DINOv3 recipes, Stage~1 groups all $14^2=196$ coarse cells together (\texttt{block\_chunk\_size=196}); each contains $4\times4$ fine queries, so one group contains all $3,136$ queries per image. Stage~2 also processes all $3,136$ sampled queries in one group (\texttt{query\_chunk\_size=3136}). Query activation checkpointing is disabled. Image and backbone features are cached once per view and reused for its queries, and each image microbatch requires one backward pass.

DINOv3-S+ uses image batches of $24$ in Stage~1 and $16$ in Stage~2, with one optimizer update per batch. DINOv3-B uses microbatches of $12$ and $8$, respectively, accumulating gradients over two microbatches to obtain the same effective batches. Each microbatch's mean loss is divided by the accumulation factor before backpropagation. Gradients are cleared at the start of each accumulation cycle, clipped to global norm $1.0$ after accumulation, and then applied in one optimizer update; the learning-rate scheduler advances once per optimizer update.

\subsection{Dense Prediction and Feature Preservation}

\begin{table}[t]
\centering
\footnotesize
\setlength{\tabcolsep}{3pt}
\caption{Task-specific probing schedules. All runs use effective batch size $4$, initial learning rate $2\times10^{-3}$, cosine decay, and final-epoch evaluation. Depth uses $256$ prediction bins rather than semantic classes.}
\label{tab:supp_probing_details}
\begin{tabular}{lccc}
\toprule
Dataset & Task & Head outputs & Epochs \\
\midrule
VOC 2012 & Segmentation & $21$ & $20$ \\
COCO-Stuff & Segmentation & $27$ & $5$ \\
ADE20K & Segmentation & $151$ & $20$ \\
Cityscapes & Segmentation & $19$ & $20$ \\
COCO & Pseudo-depth & $256$ & $5$ \\
\bottomrule
\end{tabular}
\end{table}

\paragraph{Datasets and preprocessing}
Table~\ref{tab:supp_probing_details} lists the dataset-specific schedules. Heads are trained on each dataset's training split and evaluated on its validation split. Semantic segmentation uses $21$ classes for VOC 2012, $27$ coarse classes for COCO-Stuff, $151$ labels for ADE20K (including label zero), and $19$ classes for Cityscapes. Pixels assigned ignore label $255$ are excluded. Images are resized by their shorter side and center-cropped to $448\times448$; labels use nearest-neighbor resizing. Training additionally applies horizontal flipping with probability $0.5$.

\paragraph{Task-specific heads}
All backbone and upsampler parameters are fixed. Segmentation uses a single $1\times1$ convolution from the feature dimension to the number of classes, trained with cross-entropy. Each method receives a separately trained head. Segmentation logits are bilinearly resized to $448^2$ for supervision and evaluation. All heads use AdamW with initial learning rate $2\times10^{-3}$, $(\beta_1,\beta_2)=(0.9,0.999)$, weight decay $10^{-5}$, and cosine decay over the complete training schedule. The effective batch size is $4$, the evaluation batch size is $1$, and the seed is $42$. JAFAR depth probing on COCO uses microbatches of $2$ and two-step accumulation to retain the same effective batch size. Heads are evaluated after the final training epoch. Segmentation metrics are mean IoU and pixel accuracy over valid labels.

\paragraph{Depth probing}
COCO depth targets are generated by the frozen Depth Anything V2 Small model. We concatenate each spatial feature with the broadcast, L2-normalized class token of its backbone and apply a $1\times1$ convolution from $2C$ channels to $256$ depth-bin scores. ViT-Up uses the class token from its own backbone; the other adapters use the corresponding pretrained backbone. Bin centers are uniformly spaced from $10^{-3}$ to $10$. We bilinearly resize the scores to $448^2$, then apply ReLU, add $0.1$, normalize across bins, and compute depth as the weighted mean of the bin centers. We optimize a scale-invariant logarithmic loss and multiscale gradient loss with weights $1$ and $0.5$; the logarithmic loss uses a $100$-iteration warmup. The depth teacher receives RGB values before backbone normalization. Predictions and pseudo-targets are compared on the target grid, with valid target depths restricted to $(10^{-3},10)$. We report RMSE and $\delta_1$, the fraction satisfying $\max(d/\hat d,\hat d/d)<1.25$; the main tables multiply both reported values by $100$.

\paragraph{Feature preservation}
Frozen probing reuses a dataset-specific head trained on native low-resolution DINOv3-S+ features. Neither the head nor the feature extractor is optimized during this evaluation. The same head is applied to each upsampler, with segmentation logits or depth-bin scores bilinearly resized to $448^2$ before scoring or depth decoding, respectively. This protocol is separate from training a new head on each method's features. The standard ViT-Up and pretrained-baseline dense-prediction evaluations use $448^2$ output features; in the $112^2$ comparison, ViT-Up is queried directly on the smaller grid and the head outputs are bilinearly resized to $448^2$ for supervision and scoring.

\subsection{Correspondence Evaluation}

\paragraph{SPair-71k}
We use the test split and all $18$ categories. Each image is padded to a square before resizing to $448\times448$, preserving its full field of view. Both the main comparison and the output-resolution sweep bilinearly resize source and target feature maps to $448\times448$ with \texttt{align\_corners=False}. For each annotated source keypoint, the nearest center on the resized source grid supplies the query descriptor, which is matched to the target location with maximum cosine similarity. Features are L2-normalized in FP32 after interpolation. Pixel-center coordinate transforms are applied consistently when mapping between the original image and feature grid. A match is correct when its error in original target-image coordinates is at most $\alpha$ times the longer side of the target bounding box, for $\alpha\in\{0.10,0.05,0.01\}$. Correct keypoints are aggregated within each category and the reported PCK is the unweighted mean over categories. Matching chunks contain $256$ source queries and up to $16,384$ target features; these affect memory use, not the matching criterion. Across the $481$ distinct test target boxes, the median longer side on the $448^2$ canvas is $292.1$ pixels. The corresponding median PCK tolerances are $29.2$, $14.6$, and $2.9$ pixels at thresholds $0.10$, $0.05$, and $0.01$, respectively.

\paragraph{NAVI}
We use the test split with bounding-box cropping, instance stride $4$, batch size $4$, and seed $8$. Inputs are $448\times448$ and features are evaluated on a $112\times112$ grid, with geometric grids downsampled by $0.25$. Correspondences are ranked using the nearest/second-nearest cosine-distance ratio, retaining up to $1,000$ matches per pair. Valid 3D points are transformed using the known relative camera pose. We report 3D correspondence recall at distance thresholds of $0.01$, $0.02$, and $0.05$ meters. No correspondence-specific head is trained for either benchmark.

\subsection{Ablations and Resolution Studies}

\begin{table*}[t]
\centering
\small
\renewcommand{\arraystretch}{1.12}
\caption{Settings for the reported ablations. Unlisted settings follow the corresponding Stage 1 or Stage 2 base model.}
\label{tab:supp_ablation_details}
\begin{tabular}{p{0.25\textwidth} p{0.67\textwidth}}
\toprule
Variant & Change from the base \\
\midrule
Cross-attention / FeatX off & Remove the corresponding branch from every refinement block. \\
LoRA off & Keep the pretrained backbone frozen and remove its adapters. \\
Decoder off & Remove the per-layer output decoder MLPs. \\
Transition off & Replace each transition MLP with LayerNorm, retaining normalization. \\
Learned query embedding & Replace the image-conditioned query embedding with a learned $384$-dimensional vector; disable the direct layer-$0$ loss. \\
Learned query + layer-0 conditioning & Add an initial block with layer-$0$ FeatX and fusion, without transition or cross-attention; retain layer-$0$ supervision. \\
Parallel blocks & Feed the initial query to all six blocks, sum their outputs, and use one residual $384\rightarrow1536\rightarrow384$ decoder; supervise the final output. \\
Last state repeated & Supply backbone layer $12$ to every block and supervise only the final output. \\
Last state + intermediate supervision & Supply layer $12$ to every block while retaining the base's intermediate-layer targets. \\
Shared Block Weights & Share one block across all six refinement steps while retaining independent output decoders. \\
Refinement depth & Use $T\in\{1,2,3,6,12\}$ blocks with layer indices $l[t]=12t/T$, including $l[0]=0$. \\
Stage 2 without anchor & Set the unshifted-anchor weight to $0$. \\
Stage 2 without offset regularization & Set the squared-offset penalty weights to $0$; retain shifted and anchor matching. \\
\bottomrule
\end{tabular}
\end{table*}

\begin{table}[t]
\centering
\footnotesize
\caption{Optimizer-step budgets for the convergence study. Each budget uses its own cosine schedule, with fixed effective batch size $24$ and fixed warmup length.}
\label{tab:supp_convergence_details}
\begin{tabular}{lrrr}
\toprule
Epoch budget & Total steps & Warmup & Cosine steps \\
\midrule
$1/8$ & $6,673$ & $2,669$ & $4,004$ \\
$1/4$ & $13,346$ & $2,669$ & $10,677$ \\
$1/2$ & $26,691$ & $2,669$ & $24,022$ \\
$1$ & $53,381$ & $2,669$ & $50,712$ \\
$2$ & $106,762$ & $2,669$ & $104,093$ \\
$4$ & $213,524$ & $2,669$ & $210,855$ \\
\bottomrule
\end{tabular}
\end{table}

\paragraph{Architecture and loss ablations}
The Stage~1 architecture ablations use the DINOv3-S+ training setup with padded supervision, the two feature losses, rank-$16$ LoRA, and the one-epoch schedule. Table~\ref{tab:supp_ablation_details} specifies the changes that are not fully captured by the variant names. In the final-state ablation with intermediate supervision, $l[t]$ continues to index the supervision target even though the block receives the final hidden state. Refinement-depth experiments select evenly spaced layers between embedding layer $0$ and backbone layer $12$ and adjust the decoders and supervised layers accordingly. Stage~2 ablations retain the full refinement setup and change only the indicated anchor or offset-regularization weight. VOC and Cityscapes heads are trained using the probing protocol above, and SPair-71k uses the same correspondence evaluator.

\paragraph{Training budget}
Convergence runs use separate cosine schedules for each budget in Table~\ref{tab:supp_convergence_details}, rather than truncating a longer run. The effective batch size remains $24$, and the warmup remains $2,669$ optimizer steps. The ViT-Up variants without LoRA retain the remaining settings. Each run is evaluated from its checkpoint selected by validation feature loss.

\paragraph{Output resolution}
The output-resolution sweep queries grids of $28^2$, $56^2$, $112^2$, $224^2$, and $448^2$ without retraining the upsampler. Inputs and segmentation labels remain $448^2$. VOC and Cityscapes use their frozen native-backbone heads. The separate $112^2$ versus $448^2$ dense-prediction comparison trains task-specific heads at the corresponding feature resolution. Both frozen-head and trained-head segmentation probing bilinearly resize logits to the fixed $448^2$ label grid. Depth probing likewise resizes depth-bin scores to $448^2$ before decoding. For SPair-71k, both source and target maps are decoded at the selected resolution and bilinearly resized to $448^2$ before query extraction and matching. This keeps the source-sampling and target-matching grids fixed throughout the sweep.

\paragraph{Backbone resolution}
For the joint-resolution model, we initialize from the selected DINOv3-S+ Stage~1 checkpoint and train on both $448^2$ and $896^2$ student backbone inputs. Teacher targets and query coordinates are shared between the two forwards, and their losses receive equal weight. Backbone LoRA remains trainable. This refinement uses $20,000$ optimizer steps, $1,000$ warmup steps, learning rate $2\times10^{-4}$, and microbatch size $8$ with accumulation $2$. Validation runs every $1,000$ training microbatches and averages the two resolutions. The $896^2$ evaluation uses a $56^2$ native backbone grid. In the backbone-resolution study, ViT-Up decodes $448^2$ features on both backbone grids. The ViT-Up$^{112^2}$ + Bilinear variant decodes $112^2$ instead. For native-backbone references, input sizes $448$, $896$, and $1792$ correspond to token-grid sides $28$, $56$, and $112$, respectively. Segmentation heads operate on native or ViT-Up's decoded features, with logits bilinearly resized to the common $448^2$ target grid for training and evaluation. SPair-71k uses features bilinearly resized to $448^2$ before matching.

\subsection{DPT Baseline}
\label{app:dpt_baseline}

\paragraph{Architecture}
We retain DPT's four-scale reassembly and residual fusion design~\cite{21_dpt}, adapted to dense features without class-token readout. DINOv3-S+ layers $3,6,9,12$ are reassembled at relative scales $(4,2,1,\tfrac12)$ with widths $(96,192,384,768)$ and fused using $256$ channels. The final fusion block bilinearly resizes its $112\times112$ features to $448\times448$ before its $1\times1$ projection. The unused skip branch at the coarsest level is omitted. A $3\times3$ convolution, ReLU, and $1\times1$ output projection then produce $384$-channel features on this fixed $448\times448$ grid. Resizing before these convolutions adds computation without adding parameters. The decoder has approximately $18.7$M parameters, excluding the backbone, providing capacity comparable to ViT-Up ($24.0$M).

\paragraph{Training}
We train the decoder from scratch with ViT-Up's Stage~1 padded multiscale protocol, keeping DINOv3-S+ frozen without LoRA. Bilinear grid interpolation samples the fixed output map at the training query coordinates, and average pooling aligns these predictions with the frozen teacher's final-layer features at $14^2$, $28^2$, and $56^2$. We use the same ImageNet-1K data, augmentation, and one-epoch optimization schedule as Stage~1 (Table~\ref{tab:supp_training_details}), supervising only the final decoded features with L2 and cosine losses.

\paragraph{Evaluation}
DPT produces a fixed $448^2$ feature map. Task heads and SPair-71k use this full-resolution map, while NAVI bilinearly samples the common $112^2$ matching grid from it. All requested coordinates are sampled together, and the full decoder map is computed once per image. Runtime is therefore reported only at $448^2$. 

\subsection{Baselines and Runtime Measurements}

\paragraph{Baseline implementations}
We use the supplied pretrained JAFAR, AnyUp, NAF, and UpLiFT adapters with DINOv3-S+ features. AnyUp enables NATTEN; NAF uses its pretrained model; and UpLiFT uses its optimized implementation with recursion levels selected for the requested output size. DINOv3-B comparisons use the corresponding backbone and NAF adapters, together with an UpLiFT model trained from scratch using the authors' official implementation. We train this UpLiFT model on ImageNet-1K for one epoch with $448^2$ images, batch size $32$, Adam with learning rate $10^{-3}$, and L2 feature loss. Its configuration uses two downsampling levels and iterative multi-step training, with the DINOv3-B backbone frozen. All methods follow the same dataset preprocessing and task-head protocols described above.

\paragraph{Timing and memory}
Each runtime measurement uses one NVIDIA H100 SXM GPU with $80$~GB memory and a verified $700$~W power limit, batch size $1$, and one CPU thread. Inputs are $448^2$ except in the backbone-resolution comparison. BF16 autocast is used while weights, input images, query coordinates, and returned feature tensors remain FP32. LoRA weights are merged for inference. Measurements include backbone extraction and materialization of the complete output feature map. The bilinear baseline timings in main Table~\ref{tab:ablation_backbone_resolution} are measured in a separate run on another H100 SXM under the same protocol, including feature interpolation to $448^2$. The ViT-Up$^{112^2}$ + Bilinear timings come from a further paired H100 SXM run, decoding all $112^2$ queries together and bilinearly resizing features to $448^2$. The plain ViT-Up rows retain the original direct-$448^2$ measurements. Main Table~\ref{tab:runtime_memory} instead measures directly queried output grids.

Each of five repeats performs at least $100$ warmup iterations and $3$ seconds of warmup, followed by at least $100$ measured iterations and $1$ second of measurement. The reported runtime is the median of the five repeat medians. Peak allocated CUDA memory includes the resident model, inputs, and output tensors, and is the maximum across measured iterations. ViT-Up query-chunk sizes are $112^2$, $224^2$, and $448^2$ where applicable, independently of the requested output resolution. The benchmark also records the relative spread of repeat medians, using $0.1$ as the stability tolerance.

LoftUp runtime measurements use PyTorch scaled dot-product attention (SDPA) at every output resolution, with unused attention-weight outputs disabled. These measurements are collected together on a separate H100 SXM under the same protocol.

\paragraph{Parameter accounting}
Main Tables~\ref{tab:ablation} and~\ref{tab:runtime_memory} count learned weights required for final feature prediction. This includes the final output head and LoRA before merging, but excludes the pretrained backbone, intermediate supervision heads, and Stage~2 offset head. The six-block DINOv3-S+ model has $23{,}382{,}144$ decoder weights on this basis, plus $608{,}256$ LoRA weights, totaling $23{,}990{,}400$ ($23.99$M, or $24.0$M at one decimal). The final output head contributes $148{,}608$ parameters and is retained.

Training also includes six intermediate output heads totaling $891{,}648$ parameters, giving $24{,}882{,}048$ trainable weights in Stage~1. The recorded runtime implementation merges LoRA into the backbone but retains these intermediate heads and the inactive $99{,}842$-parameter offset head. It therefore holds $24{,}373{,}634$ non-backbone parameters. Their storage and any computation retained by that implementation are included in the reported runtime and memory measurements. The offset head is not executed during feature inference.

\section{Video Object Segmentation}
\label{sec:supp_video_segmentation}

\begin{table*}[t]
\centering
\footnotesize
\setlength{\tabcolsep}{5pt}
\caption{DAVIS-2017 val video object segmentation with linear probing (\%). Grid is the upsampler's backbone token grid on the square input canvas; native is the backbone at the frame's own resolution. Features use the method's dense output alone (dense), LILA's sum with the bilinear backbone branch (comb.), or the backbone branch alone (backbone). ViT-Up (joint) is trained on two backbone grids. $\mathcal{J}$ and $\mathcal{F}$ report mean (M), recall (R), and decay (D; lower is better). Probing $\mathcal{J}\&\mathcal{F}$ is the mean $\pm$ standard deviation over $10$ seeds; the other metrics are seed means.}
\label{tab:supp_video_probe}
\begin{tabular}{llc c ccc ccc}
\toprule
\textbf{Method} & \textbf{Grid} & \textbf{Features} & $\mathcal{J}\&\mathcal{F}\!\uparrow$ & $\mathcal{J}_\mathrm{M}\!\uparrow$ & $\mathcal{J}_\mathrm{R}\!\uparrow$ & $\mathcal{J}_\mathrm{D}\!\downarrow$ & $\mathcal{F}_\mathrm{M}\!\uparrow$ & $\mathcal{F}_\mathrm{R}\!\uparrow$ & $\mathcal{F}_\mathrm{D}\!\downarrow$ \\
\midrule
\multicolumn{10}{l}{\textit{DINOv2-S}} \\
Bilinear & native & backbone & 58.6\,{\scriptsize$\pm$0.5} & 55.7 & 65.0 & 25.9 & 61.6 & 69.3 & 30.2 \\
LILA & native & dense & 64.5\,{\scriptsize$\pm$0.2} & 62.0 & 70.8 & 25.0 & 67.0 & 75.0 & 25.5 \\
 &  & comb. & 66.2\,{\scriptsize$\pm$0.6} & 63.2 & 71.3 & 25.6 & 69.2 & 76.5 & 28.5 \\
NAF & $60^2$ & dense & 62.6\,{\scriptsize$\pm$0.3} & 60.4 & 69.8 & 24.3 & 64.8 & 73.3 & 26.2 \\
 &  & comb. & 64.2\,{\scriptsize$\pm$0.4} & 61.2 & 69.7 & 26.6 & 67.2 & 75.2 & 30.4 \\
UpLiFT & $60^2$ & dense & 65.1\,{\scriptsize$\pm$0.4} & 62.6 & 71.6 & 25.1 & 67.5 & 75.4 & 26.6 \\
 &  & comb. & 64.1\,{\scriptsize$\pm$0.6} & 61.4 & 69.6 & 26.2 & 66.9 & 74.5 & 29.5 \\
ViT-Up & $32^2$ & dense & 64.8\,{\scriptsize$\pm$0.2} & 62.3 & 70.8 & 23.1 & 67.4 & 75.8 & 26.4 \\
 &  & comb. & 65.9\,{\scriptsize$\pm$0.5} & 62.7 & 71.2 & 26.8 & 69.1 & 76.6 & 29.8 \\
ViT-Up (joint) & $60^2$ & dense & 67.3\,{\scriptsize$\pm$0.2} & 64.7 & 71.7 & 25.0 & 69.9 & 77.8 & 26.6 \\
 &  & comb. & 66.3\,{\scriptsize$\pm$0.6} & 63.0 & 70.5 & 27.2 & 69.5 & 76.4 & 29.2 \\
\midrule
\multicolumn{10}{l}{\textit{DINOv3-S+}} \\
Bilinear & native & backbone & 63.7\,{\scriptsize$\pm$0.2} & 60.7 & 69.8 & 21.6 & 66.7 & 75.1 & 24.1 \\
NAF & $28^2$ & dense & 55.9\,{\scriptsize$\pm$0.3} & 54.3 & 62.0 & 22.7 & 57.5 & 66.1 & 23.9 \\
 &  & comb. & 62.1\,{\scriptsize$\pm$0.5} & 59.9 & 67.7 & 24.9 & 64.4 & 71.0 & 27.2 \\
UpLiFT & $28^2$ & dense & 54.9\,{\scriptsize$\pm$0.4} & 53.6 & 61.0 & 24.5 & 56.1 & 63.0 & 26.3 \\
 &  & comb. & 61.3\,{\scriptsize$\pm$0.3} & 58.8 & 66.9 & 25.1 & 63.8 & 70.4 & 27.6 \\
ViT-Up & $28^2$ & dense & 66.0\,{\scriptsize$\pm$0.4} & 63.5 & 71.7 & 23.8 & 68.5 & 75.7 & 25.9 \\
 &  & comb. & 67.5\,{\scriptsize$\pm$0.3} & 64.8 & 73.2 & 23.7 & 70.2 & 78.1 & 25.8 \\
NAF & $56^2$ & dense & 63.9\,{\scriptsize$\pm$0.3} & 61.5 & 68.9 & 22.0 & 66.3 & 76.3 & 24.3 \\
 &  & comb. & 66.3\,{\scriptsize$\pm$0.3} & 63.8 & 70.9 & 23.5 & 68.9 & 77.3 & 26.2 \\
UpLiFT & $56^2$ & dense & 66.0\,{\scriptsize$\pm$0.3} & 63.6 & 71.8 & 23.8 & 68.4 & 77.2 & 25.1 \\
 &  & comb. & 67.1\,{\scriptsize$\pm$0.4} & 64.5 & 72.3 & 23.9 & 69.8 & 78.5 & 24.9 \\
ViT-Up (joint) & $56^2$ & dense & 69.5\,{\scriptsize$\pm$0.3} & 66.8 & 73.8 & 24.1 & 72.2 & 79.6 & 26.2 \\
 &  & comb. & 70.1\,{\scriptsize$\pm$0.1} & 67.3 & 74.5 & 23.6 & 72.8 & 80.6 & 25.3 \\
\bottomrule
\end{tabular}
\end{table*}

\begin{table*}[t]
\centering
\footnotesize
\setlength{\tabcolsep}{5pt}
\caption{DAVIS-2017 val video object segmentation with local k-NN propagation (\%). Grid is the upsampler's backbone token grid on the square input canvas; native is the backbone at the frame's own resolution. Features use the method's dense output alone (dense), LILA's sum with the bilinear backbone branch (comb.), or the backbone branch alone (backbone). ViT-Up (joint) is trained on two backbone grids. $\mathcal{J}$ and $\mathcal{F}$ report mean (M), recall (R), and decay (D; lower is better).}
\label{tab:supp_video_knn}
\begin{tabular}{llc c ccc ccc}
\toprule
\textbf{Method} & \textbf{Grid} & \textbf{Features} & $\mathcal{J}\&\mathcal{F}\!\uparrow$ & $\mathcal{J}_\mathrm{M}\!\uparrow$ & $\mathcal{J}_\mathrm{R}\!\uparrow$ & $\mathcal{J}_\mathrm{D}\!\downarrow$ & $\mathcal{F}_\mathrm{M}\!\uparrow$ & $\mathcal{F}_\mathrm{R}\!\uparrow$ & $\mathcal{F}_\mathrm{D}\!\downarrow$ \\
\midrule
\multicolumn{10}{l}{\textit{DINOv2-S}} \\
Bilinear & native & backbone & 62.2 & 61.7 & 76.5 & 18.2 & 62.7 & 74.4 & 27.8 \\
LILA & native & dense & 73.9 & 71.7 & 85.7 & 11.5 & 76.0 & 87.1 & 16.8 \\
 &  & comb. & 68.5 & 67.3 & 80.3 & 18.2 & 69.8 & 80.7 & 25.7 \\
NAF & $60^2$ & dense & 65.9 & 64.4 & 77.8 & 18.0 & 67.3 & 80.5 & 22.4 \\
 &  & comb. & 62.2 & 61.4 & 74.2 & 24.0 & 63.1 & 72.3 & 31.8 \\
UpLiFT & $60^2$ & dense & 69.9 & 67.6 & 80.1 & 17.7 & 72.3 & 83.3 & 21.5 \\
 &  & comb. & 64.9 & 63.4 & 77.1 & 22.8 & 66.3 & 77.7 & 29.5 \\
ViT-Up & $32^2$ & dense & 71.7 & 68.9 & 80.2 & 14.9 & 74.5 & 84.7 & 19.3 \\
 &  & comb. & 66.2 & 64.3 & 77.1 & 22.3 & 68.1 & 78.6 & 28.8 \\
ViT-Up (joint) & $60^2$ & dense & 76.0 & 72.5 & 84.8 & 12.4 & 79.6 & 88.2 & 14.5 \\
 &  & comb. & 69.1 & 67.1 & 80.5 & 18.6 & 71.1 & 82.0 & 25.6 \\
\midrule
\multicolumn{10}{l}{\textit{DINOv3-S+}} \\
Bilinear & native & backbone & 63.3 & 61.8 & 73.7 & 18.6 & 64.8 & 76.4 & 26.2 \\
NAF & $28^2$ & dense & 58.1 & 58.2 & 68.9 & 20.0 & 58.1 & 69.3 & 26.4 \\
 &  & comb. & 57.3 & 57.5 & 65.8 & 26.2 & 57.2 & 65.7 & 33.6 \\
UpLiFT & $28^2$ & dense & 55.6 & 55.7 & 64.5 & 22.3 & 55.6 & 64.9 & 28.3 \\
 &  & comb. & 57.5 & 57.6 & 65.3 & 24.8 & 57.4 & 64.7 & 32.0 \\
ViT-Up & $28^2$ & dense & 69.5 & 67.1 & 78.1 & 18.7 & 71.9 & 80.8 & 23.3 \\
 &  & comb. & 64.8 & 63.6 & 74.0 & 22.3 & 66.0 & 76.4 & 28.6 \\
NAF & $56^2$ & dense & 62.8 & 61.1 & 71.3 & 21.2 & 64.5 & 75.4 & 26.2 \\
 &  & comb. & 59.9 & 59.3 & 68.2 & 25.1 & 60.4 & 69.0 & 32.7 \\
UpLiFT & $56^2$ & dense & 67.4 & 65.7 & 77.5 & 18.5 & 69.1 & 79.7 & 24.0 \\
 &  & comb. & 63.8 & 62.8 & 72.5 & 21.2 & 64.7 & 74.7 & 27.7 \\
ViT-Up (joint) & $56^2$ & dense & 73.1 & 70.3 & 80.6 & 16.6 & 75.9 & 85.4 & 21.2 \\
 &  & comb. & 66.3 & 64.8 & 76.1 & 21.5 & 67.8 & 77.8 & 27.3 \\
\bottomrule
\end{tabular}
\end{table*}

\subsection{Protocol}

We port LILA's DAVIS-2017 evaluation code~\cite{26_lila} and evaluate all $30$ validation videos. Following the official DAVIS evaluation~\cite{17_davis}, the first and last frames are excluded from $\mathcal{J}$ and $\mathcal{F}$. All methods, including LILA's released DINOv2-S checkpoint, are evaluated with the same code.

\paragraph{Linear probing} Frames are resized to a shorter side of $476$ pixels ($480$ for the $16$-pixel patches of DINOv3-S+), with a patch-aligned longer side. For each video, a $1{\times}1$ convolution is fitted to the features of the first frame and its annotated mask for $2{,}000$ AdamW steps with learning rate $5\times10^{-4}$ and weight decay $10^{-4}$. Logits are bilinearly resized to the label resolution. The fitted probe then segments every frame independently, without temporal context. We train $10$ probes with different random initializations and report the mean and standard deviation.

\paragraph{Local k-NN propagation} Frames are resized to a shorter side of $476$ pixels with a $14$-pixel-aligned longer side for all backbones, as in LILA's code. Features are bilinearly resized to a grid with shorter side $60$, $\ell_2$-normalized, and scaled by temperature $0.1$. Each pixel's label is propagated from its five nearest neighbors within a $25{\times}25$ window over the first frame and the last seven predicted frames~\cite{20_crw,21_dino}. Predicted masks are bilinearly upsampled to the frame resolution.

\subsection{Features and Resolution}

Every probe and propagation step uses the frozen backbone's final patch tokens (the encoder branch) and the method's dense features (the decoder branch). Dense features are layer-normalized without affine parameters, mirroring LILA's normalized decoder output. We evaluate three feature modes in one pass: the backbone branch alone, the dense features alone, and LILA's combined protocol. The backbone branch alone is the bilinear baseline: probing bilinearly resizes its logits, and k-NN resizes its tokens. In combined probing, the logits of both branches are summed with unit weights. In combined k-NN, LILA's released code sums the soft masks of both branches and feeds the sum back as temporal context, so the masks double every frame and overflow in half precision after approximately $16$ frames ($36.6$ $\mathcal{J}\&\mathcal{F}$ for LILA's checkpoint). We average the two soft masks instead.

Each method produces dense features at the requested resolution: the probe's input resolution or the k-NN grid. All upsamplers operate on square inputs: frames are centered on a padded black square canvas resized to the upsampler's backbone input. ViT-Up queries the cell centers of the requested grid within the frame. NAF and UpLiFT compute their native output, which is cropped to the frame and bilinearly resized. LILA's decoder reads the encoder tokens at the frame's own aspect ratio. In Table~\ref{tab:video_segmentation}, DINOv2-S upsamplers use an $840^2$ canvas ($60^2$ tokens), on which a DAVIS frame covers $34{\times}60$ tokens, the same as LILA's encoder on $476{\times}840$ frames. DINOv3-S+ upsamplers use an $896^2$ canvas ($56^2$ tokens).

\subsection{Joint-Grid Models}

The joint-resolution ViT-Up models used for video segmentation extend the joint-resolution Stage~1 refinement to Stage~2. Stage~2 initializes from the best joint Stage~1 checkpoint and trains on both $448^2$ and $896^2$ student backbone inputs, with teacher inputs of the same size as each student input. It uses $20{,}000$ optimizer steps, $1{,}000$ warmup steps, learning rate $2\times10^{-4}$, and microbatch size $8$ with accumulation $2$. For DINOv2-S, both stages train on $32^2$ and $64^2$ grids, starting from the DINOv2-S Stage~1 model. We select the best checkpoints by validation loss (DINOv3-S+ and DINOv2-S Stage~2: step $19{,}000$). The DINOv2-S model was not trained on $60^2$ grids; at $840^2$, it is evaluated between its two training resolutions.

\subsection{Comparison with Reported Numbers}

LILA reports $57.5$ probing and $65.1$ k-NN $\mathcal{J}\&\mathcal{F}$ for the DINOv2-S backbone, and $68.6$ and $73.9$ with LILA~\cite{26_lila}. Our backbone probing result ($58.6$) is close to the reported value. LILA states that its k-NN evaluation follows DINO's settings; its released k-NN code gives $62.2$ for the backbone, which we use for every method. For LILA's checkpoint, our dense k-NN result reproduces the reported $73.9$, whereas the combined result is $68.5$. Our probing results are $64.5$ (dense) and $66.2$ (combined), below the reported $68.6$. Because the reported numbers do not specify the feature mode and the k-NN baseline could not be reproduced, we compare methods only with numbers from our common evaluation.

\subsection{Additional Results}

Tables~\ref{tab:supp_video_probe} and~\ref{tab:supp_video_knn} report all feature modes, grids, and DAVIS metrics. The single-grid ViT-Up models ($32^2$ for DINOv2-S and $28^2$ for DINOv3-S+) already outperform NAF and UpLiFT at the same or larger grids in dense k-NN propagation. In probing, the $28^2$ DINOv3-S+ model matches UpLiFT at $56^2$, while the $32^2$ DINOv2-S model remains below UpLiFT at $60^2$. NAF and UpLiFT fall below the bilinear baseline at $28^2$ on DINOv3-S+, where a frame covers only $16{\times}28$ tokens on the square canvas, fewer than the backbone branch's native tokens.

\end{document}